\PassOptionsToPackage{table,xcdraw}{xcolor}
\documentclass[preprint,11pt]{elsarticle}

\usepackage[a4paper, total={5.5in, 10in}]{geometry}

\usepackage{graphicx} 
\usepackage{subcaption} 

\usepackage{booktabs}
\usepackage{longtable}
\usepackage{tabularx} 
\usepackage{multirow}

\usepackage{enumitem}

\usepackage{amsmath, amssymb, amsfonts, amsthm}

\usepackage{pgfplots}
\usepackage{tikz}
\usetikzlibrary{shapes.geometric, arrows, positioning}

\usepackage{hyperref}
\hypersetup{
    colorlinks=true,
    linkcolor=blue,
    citecolor=blue,
    urlcolor=blue
}

\usepackage{enumitem}
\usepackage{multirow}
\usepackage{pdflscape} 
\usepackage{enumitem}

\usepackage{array}
\renewcommand{\arraystretch}{1.2}

\begin{document}

\begin{frontmatter}



\title{Information Fusion in Smart Agriculture: Machine Learning Applications and Future Research Directions}


\author[mfs]{Aashu Katharria}
\ead{aashu@mfs.iitr.ac.in}

\author[ISI]{Kanchan Rajwar}
\ead{krajwar@ma.iitr.ac.in}

\author[mfs,ie]{Millie Pant} \corref{cor1}
\ead{pant.milli@as.iitr.ac.in}

\author[apm,isci]{Juan D. Velásquez}
\ead{jvelasqu@dii.uchile.cl}

\author[cs]{Václav Snášel}
\ead{vaclav.snasel@vsb.cz}

\author[math]{Kusum Deep}
\ead{kusum.deep@ma.iitr.ac.in}


\affiliation[mfs]{organization={Mehta Family School of Data Science and Artificial Intelligence, Indian Institute of Technology Roorkee}, 
            city={Roorkee}, 
            country={India}}

\affiliation[ISI]{organization={Statistical Quality Control and Operations Research Unit, Indian Statistical Institute}, 
            city={Hyderabad}, 
            country={India}}
            
\affiliation[ie]{organization={Department of Applied Mathematics and Scientific Computing, Indian Institute of Technology Roorkee}, 
            city={Roorkee}, 
            country={India}}

\affiliation[apm]{organization={Department of Industrial Engineering, University of Chile}, 
            city={Santiago}, 
            country={Chile}}

\affiliation[isci]{organization={Instituto Sistemas Complejos de Ingeniería}, 
            city={Santiago}, 
            country={Chile}}
\affiliation[cs]{organization={Department of Computer Science, VSB-Technical University of Ostrava}, 
            city={Ostrava}, 
            country={Czech Republic}}

\affiliation[math]{organization={Department of Mathematics, Indian Institute of Technology Roorkee}, 
            city={Roorkee}, 
            country={India}}


\begin{abstract}

Machine learning (ML) is a rapidly evolving technology with expanding applications across various fields. This paper presents a comprehensive survey of recent ML applications in agriculture for sustainability and efficiency. Existing reviews mainly focus on narrow subdomains or lack a fusion-driven perspectives. This study provides a combined analysis of ML applications in agriculture, structured around five key objectives: (i) Analyzing ML techniques across pre-harvesting, harvesting, and post-harvesting phases. (ii) Demonstrating how ML can be used with agricultural data and data fusion. (iii) Conducting a bibliometric and statistical analysis to reveal research trends and activity. (iv) Investigating real-world case studies of leading artificial intelligence (AI)-driven agricultural companies that use different types of multisensors and multisource data.  (v) Compiling publicly available datasets to support ML model training. Going beyond existing previous reviews, 
this review focuses on how machine learning (ML) techniques, combined with multi-source data fusion (integrating remote sensing, IoT, and climate analytics), enhance precision agriculture by improving predictive accuracy and decision-making. Case studies and statistical insights illustrate the evolving landscape of AI driven smart farming, while future research directions also discusses challenges associated with data fusion for heterogeneous datasets. This review bridges the gap between AI research and agricultural applications, offering a roadmap for researchers, industry professionals, and policymakers to harness information fusion and ML for advancing precision agriculture.

\end{abstract}


\begin{highlights}
\item Comprehensive review of ML applications in pre-harvest, harvest, and post-harvest.
    \item Emphasizes multi-source data fusion for improved precision farming and decision-making.
    \item Includes bibliometric and statistical analysis of ML research trends in agriculture.
    \item Explores real-world AI applications and industry innovations in smart agriculture.
    \item Presents key datasets supporting ML model training for agricultural advancements.

\end{highlights}


\end{frontmatter}

\section{Introduction}
\label{sec 1}

Artificial intelligence (AI) is a powerful technologies with applications in various fields, such as healthcare, agriculture, cybersecurity, finance, transportation, and many more. Recently, machine learning (ML) has made significant advancements in agricultural applications, driving innovation and efficiency in the field. Agriculture encompasses a wide range of practices, including crop monitoring, weather predictions for crop yield, livestock production, aquaculture, and forestry, which provide food and non-food products such as fibers, fuels, and raw materials like rubber \cite{sunil2023systematic}. It is an essential sector for feeding the growing world population, which is predicted by the Food and Agriculture Organization (FAO) to reach approximately nine billion by 2050 \cite{talaviya2020implementation}. Agriculture also plays a critical role in the global economy, providing a foundation for sustainable growth and long-term economic development. Many countries have recognized the importance of this sector, as demonstrated by its contribution to their gross domestic product (GDP). In 2022 alone, the United States' agricultural GDP reached \$201 billion in the fourth quarter of 2022 \cite{usda_ag_food_stats}, while China \cite{statista_china_gdp}, Japan \cite{statista_japan_gdp} and India also made notable contributions, reflecting the economic weight of the sector. The increasing demand for food requires continued focus on agricultural innovation and investment \cite{jarray2023machine}. 

Throughout history, four major agricultural revolutions have transformed farming. The \textit{\textbf{First Agricultural Revolution}} (Neolithic, $\sim$10,000 BCE) introduced domestication of plants and animals. The \textit{\textbf{Second Agricultural Revolution}} (18th--19th centuries) in Europe and North America brought innovations like the plow, seed drilling, and crop rotation, boosting productivity. The \textit{\textbf{Third Agricultural Revolution}} (Green Revolution, mid-20th century) introduced high-yield crops, fertilizers, pesticides, mechanization, and irrigation. 
Now, the \textit{\textbf{fourth agricultural revolution}}; Agriculture 4.0 is underway, driven by AI, multi-sensor data fusion, and hybrid architectures, optimizing precision farming with diverse real-world datasets 
\cite{harwood2020history,  rithani2023review, liu2020industry}.

ML is an interdisciplinary field that merges computer science and statistics to perform tasks that are typically performed by humans. In recent years, the integration of ML with big data technologies and high-performance computing has gained popularity in agriculture, facilitating the analysis of data-intensive processes. Also, to enhance the accuracy of ML applications and predictions, researchers and practitioners integrate data from multiple sources, a process known as information fusion \cite{allu2025remotesensing}. ML and data fusion have become an essential tool in agriculture and have found applications in various scientific fields such as bio-informatics \cite{larranaga2006machine}, biochemistry \cite{mowbray2021machine}, medicine \cite{sidey2019machine}, meteorology \cite{chase2022machine}, economic sciences \cite{athey2018impact}, robotics \cite{pierson2017deep}, aquaculture \cite{zhao2021application}, food security \cite{sood2021computer}, and climatology \cite{bochenek2022machine}.

Agriculture, like other domains, is being transformed by ML. Agriculture 4.0 merges traditional practices with IoT, AI, and automation, utilizing data fusion from sensors, drones, and AI-driven machinery to enhance efficiency, sustainability, yields, and waste reduction. It is a direct response to growing food needs and environmental challenges, reflecting the readiness of the agricultural sector to embrace innovation and adapt to a rapidly changing world. Precision agriculture \cite{zhang2002precision}, digital agriculture \cite{basso2020digital}, and smart agriculture \cite{lipper2014climate} are terms used to describe the use of technology and data fusion in agricultural practices. These evolving agricultural practices hold great promise for improving efficiency, sustainability, and productivity in agriculture. The study \cite{mckinion1985expert} is the earliest known work in the literature to introduce AI in agriculture, where expert systems were developed as decision aids for integrated crop management, encompassing areas such as irrigation, fertilization, weed control, and pest management. These expert systems aimed to enhance decision-making in agricultural practices by using AI-based reasoning and analysis to improve efficiency and productivity.

However, the agriculture sector faces challenges that include climate change, limited resources, and the need for sustainable practices. Unpredictable weather patterns, pests, and diseases affect crop yields. Limited land, water, and energy resources strain production. ML offers innovative solutions to improve productivity and address these challenges. Although some review articles (details are provided in Section~\ref{Sec 2})  have explored various ML models, a comprehensive review is still needed in this emerging field.  Consequently, this study aims to explore the following research objective (RO):

    \begin{enumerate}[label=\textit{\textbf{RO-\arabic*:}}, leftmargin=*, labelsep=1em]
    
            \item Investigating recent advancements and applications of ML in pre-harvesting, harvesting, and post-harvesting processes, highlighting early disease detection, weather prediction, seed testing, irrigation planning, and crop recommendation.
    
            \item Demonstrating how ML can be used with agricultural data and data fusion to provide useful insights.
            
            \item Understanding current research activity and conducting a bibliometric analysis by accumulating statistical data from the Scopus directory.
            
            \item Examining real-world applications of the fusion of ML and agriculture, including case studies of the top 10 AI-based companies based on valuation that address the gap between academia and practical implementation.
            
            \item Listing popular datasets for training ML models, providing resources for practitioners.
            
    \end{enumerate}

\begin{figure*}[t]
        \centering
        \includegraphics[width=0.9\textwidth]{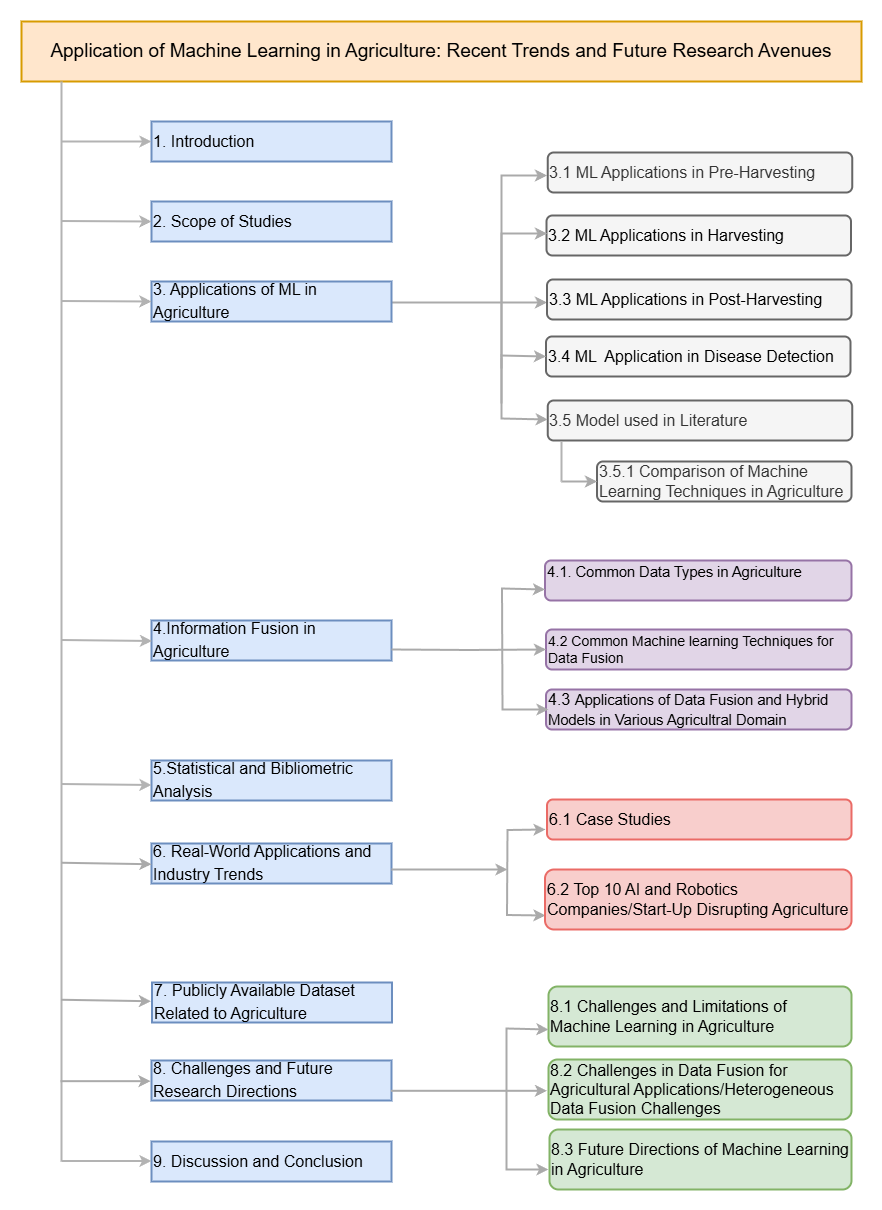}
        \caption{The organization of this paper with sections and subsections.}
        \label{paper_s}
    \end{figure*}

The structure of the paper is illustrated in Fig.~\ref{paper_s}, which outlines the sections and subsections.

\section{Scope of Studies}
\label{Sec 2}

The field of agricultural ML has seen a significant increase in research, resulting in a substantial volume of research articles. In the last two years alone, more than 5,000 articles have been published involving the keywords ``agriculture" and ``machine learning." (A detailed quantitative analysis of this trend is provided in Section \ref{Sec 5}, particularly in Fig.~\ref{research}.) Therefore, conducting a comprehensive systematic review on the ML and agriculture domain seems nearly impossible. This is why most available systematic reviews focus on specific niches within agriculture, such as soil classification \cite{srivastava2021comprehensive}, disease detection \cite{li2021plant}, or crop recommendation \cite{van2020crop} using ML. 

Although some comprehensive review articles are available on the general domain of ML and agriculture, they are not up-to-date and do not cover all aspects. A summary of some of these review articles, highlighting their focus areas and key contributions, is provided in Table \ref{review}. Therefore, by focusing on \textbf{RO-1} to \textbf{RO-5}, this study provides a meaningful up-to-date survey of the fusion between ML and agriculture. Additionally, given the critical role of multiple data sources—such as remote sensing data (satellite, aerial, drone, LiDAR, and radar data)—this study emphasizes data fusion and its impact on agricultural advancements.

Our methodology consisted of a three-stage process: identifying relevant studies, screening them, and evaluating their eligibility. This approach allowed us to outline the distinct features, advantages, and shortcomings of each application, thus highlighting both the potential benefits and the challenges inherent in the deployment of ML in agricultural settings. Relevant data has been collected from the Scopus database.

\begin{table*}[!htp]\centering
    \caption{Summary of Review Papers on Machine Learning in Agriculture.}\label{review}
    \scriptsize 
    \resizebox{\textwidth}{!}{ 
    \begin{tabular}{|p{1.9cm}|p{1.5cm}p{1.5cm}p{1.5cm}p{2.0cm}p{1.8cm}p{1.5cm}p{1.5cm}|p{1.2cm}p{1.2cm}p{1.5cm}p{1cm}|p{1.5cm}|}
        \toprule
        \midrule
        & \multicolumn{7}{c}{\cellcolor{gray!20}Analysis} & \multicolumn{4}{c}{\cellcolor{gray!40}Type of Review} & \multicolumn{1}{c}{\cellcolor{gray!60}Data Fusion} \\
        \midrule
        Ref. & Pesticide \newline Management & Weed \newline Detection & Disease \newline Detection & Crop Recommen-\newline dation System & Automatic \newline Irrigation  Control & Yield \newline Management & Soil/Seed \newline Analysis & System-\newline atic & Bibliog-\newline raphy & Compre-\newline hensive & Mixed & Data Fusion Covered \\
        \midrule
        \cite{saleem2021automation} & $\times$ & $\times$ & \checkmark & $\times$ & \checkmark & \checkmark & \checkmark & $\times$ & $\times$ & \checkmark & \checkmark & $\times$ \\
        \cite{pathan2020artificial} & $\times$ & $\times$ & \checkmark & $\times$ & $\times$ & $\times$ & $\times$ & $\times$ & $\times$ & $\times$ & \checkmark & $\times$ \\
        \cite{jha2019comprehensive} & \checkmark & \checkmark & \checkmark & $\times$ & $\times$ & $\times$ & $\times$ & $\times$ & $\times$ & \checkmark & \checkmark & $\times$ \\
        \cite{dhanya2022deep} & \checkmark & \checkmark & \checkmark & $\times$ & \checkmark & \checkmark & \checkmark & $\times$ & $\times$ & $\times$ & $\times$ & $\times$ \\
        \cite{bannerjee2018artificial} & \checkmark & \checkmark & \checkmark & $\times$ & \checkmark & \checkmark & \checkmark & $\times$ & $\times$ & $\times$ & $\times$ & $\times$ \\
        \cite{sharma2020machine} & \checkmark & \checkmark & \checkmark & $\times$ & $\times$ & $\times$ & $\times$ & \checkmark & $\times$ & $\times$ & $\times$ & $\times$ \\
        \cite{cisternas2020systematic} & $\times$ & \checkmark & $\times$ & $\times$ & \checkmark & $\times$ & \checkmark & $\times$ & $\times$ & $\times$ & $\times$ & $\times$ \\
        \cite{shaikh2022towards} & \checkmark & \checkmark & \checkmark & $\times$ & $\times$ & $\times$ & $\times$ & $\times$ & $\times$ & $\times$ & $\times$ & $\times$ \\
        This paper & \checkmark & \checkmark & \checkmark & \checkmark & \checkmark & \checkmark & \checkmark & $\times$ & \checkmark & \checkmark & \checkmark & \checkmark \\
        \midrule
        \bottomrule
    \end{tabular}
    } 
\end{table*}

However, this comprehensive review does not focus on the technical intricacies of individual ML algorithms, nor does it extensively explore the broader socio-economic impacts of agricultural AI. Instead, our objective is to present a holistic and valuable analysis that offers readers a comprehensive understanding of the current state and future potential of ML in the transformation of agriculture. 

\section{Applications of Machine Learning in Agriculture} 
\label{Sec 3}

To address \textbf{RQ-1}, the application of ML techniques in agriculture is extensively examined, highlighting their potential to enhance farming practices, optimize crop yields, and reduce losses. The effectiveness of these techniques across various stages of agriculture is demonstrated as follows:
\subsection{Machine Learning Applications in Pre-Harvesting}
\label{sec:ml_Pre_Harvesting}

Numerous studies have explored the application of ML in pre-harvesting. For example, Radhika et al. \cite{radhika2009atmospheric} compared support vector regression (SVR) and multi-layer perceptron (MLP) for atmospheric temperature prediction, highlighting the superior performance of SVR. Srunitha et al. \cite{srunitha2016performance} demonstrated the effectiveness of support vector machines (SVM) in accurately classifying non-sandy soils, suggesting the potential for ML to enhance soil property prediction. Nahvi et al. \cite{nahvi2016using} focused on improving the accuracy of estimating soil temperature using the Extreme Learning Machine (ELM) technique with the Self-Adaptive Evolutionary (SaE) algorithm. Akshay et al. \cite{akshay2020efficient} emphasized the importance of precise soil moisture measurement and management for optimal crop yield, proposing the K-Nearest Neighbors (K-NN) algorithm for data analysis. Uzal et al. \cite{uzal2018seed} found that CNNs outperformed traditional methods to estimate the number of seeds per soybean pod, demonstrating robust behavior against seasonal changes. Reddy et al. \cite{reddy2020iot} proposed an IoT-based smart irrigation system that predicts water requirements using a decision tree algorithm to improve irrigation efficiency. Ai et al. \cite{ai2020research} focused on using CNNs for automated crop disease identification, achieving high precision, and developing a WeChat applet for disease recognition. Mridha et al. \cite{mridha2021plant} proposed a deep learning-based method for early detection and identification of plant diseases, outperforming traditional models and offering a web-based AI tool. Muhammed et al. \cite{muhammed2022user} presented a user-friendly crop recommendation system (UACR) that uses AIoT architecture to assist farmers in selecting suitable crops based on location, rain level and soil type.  

\subsection{Machine Learning Applications in Harvesting}
\label{sec:ml_harvesting}

Significant advances have been made in the use of ML for harvesting purposes. Durmucs-Bunlu et al. \cite{durmucs2015design} discussed the development of a mobile autonomous robot designed for agricultural operations, with the goal of enhancing production efficiency through precision farming, disease diagnosis, and soil analysis. De Baerdemaeker et al. \cite{de2018development} introduced Octinion's fully autonomous picking robot for strawberry cultivation, addressing labor shortages in the industry. This robot effectively identifies and harvests ripe strawberries, ensuring their quality and neatly arranging them in punnets. Zhang et al. \cite{zhang2018deep} presented an improved deep learning-based method for classification of tomato ripeness in a harvesting robot. The study explored various techniques for augmentation of data sets to improve accuracy using CNN. Kestur et al. \cite{kestur2019mangonet} proposed MangoNet, a deep learning system that utilizes a convolutional neural network and contour-based object detection for mango detection and counting in RGB images, enabling the estimation of mango yield. Altaheri et al. \cite{altaheri2019date} described the development, testing and validation of SWEEPER, an autonomous robot that is used to harvest sweet pepper fruits in greenhouses. SWEEPER employs an industrial arm and advanced sensing technology to perform its harvesting tasks. Furthermore, Bogue et al. \cite{bogue2020fruit} provided information on recent commercial developments in fruit picking robots, focusing specifically on strawberry and other soft fruit harvesting, while examining their commercial prospects. 

\subsection{Machine Learning Applications in Post-Harvesting}
\label{sec:ml_Post_harvesting}

In the field of post-harvesting, ML has demonstrated its potential in various applications. Al-Hilali et al. \cite{al2011computer} address the challenges faced by Saudi Arabian growers by proposing a visionary solution that harnesses the power of computer vision and RGB images. Their groundbreaking system extracts quality features and achieves precise categorization of dates with an impressive 80\% accuracy using a back-propagation neural network. Purandare et al. \cite{purandare2016analysis} present an end-to-end system aimed at reducing post-harvest losses in agriculture. This system provides real-time data, suggestions for optimal harvesting, and predictive analytics for warehouse management. Using statistical and probabilistic techniques, it mitigates storage and transportation losses while supporting informed decision-making. Chakravarthy et al. \cite{chakravarthy2019micro} introduce an automated system to classify mangoes according to their condition using image processing techniques. Using MATLAB and the k-means clustering algorithm, this study extracts characteristics from mango images and evaluates the precision of the classification process, ultimately improving efficiency in the classification and classification of fruits. Another notable application is the tomato grading machine vision system described by Ireri et al. \cite{ireri2019computer}. This system utilizes RGB images to achieve accurate and reliable grading by detecting calyx and stalk scars, identifying defected regions, and recognizing various grading categories based on color and texture features. With its high accuracy and potential for inline sorting, this system plays a crucial role in maintaining quality standards during tomato production. Duysak et al. \cite{duysak2020machine} employ the K-nearest neighbor (KNN) algorithm and stepped frequency continuous wave radar (SFCWR) to accurately determine grain quantity in a silo. Through the analysis of back-scattering signals, range profiles, and feature combinations, they achieve an impressive $96.71\%$ accuracy, highlighting the effectiveness of ML in assessing grain quantity with precision.
Lutz et al. \cite{lutz2022applications} propose a hierarchical model called Boosted Continuous Non-spatial whole Attribute Extraction (BCNAE) for detecting post-harvest losses in sekai-ichi apples. This model comprises three levels: manual separation, shape fitting, and feature refinement through image segmentation. The BCNAE framework demonstrates effective detection and reduction of post-harvest losses, offering precise solutions in apple production. Table \ref{tab:long} provides a comprehensive overview of the various applications of ML in the field of agriculture.

\begin{landscape}
{\scriptsize     
\begin{longtable}{|p{0.5cm}|p{2cm}|p{0.25cm}|p{7.0cm}|p{2.5cm}|p{4.75cm}|p{4.75cm}|p{0.2cm}|} 
\caption{Applications of ML at Different Stages of Agriculture (Pre-harvesting, Harvesting, Post-harvesting).}
\label{tab:long}\\
\toprule
\midrule
\multicolumn{1}{|c|}{\small \textbf{Sn}} & \multicolumn{1}{|c|}{\small \textbf{Application}} & \multicolumn{1}{|c|}{\small \textbf{Year}} & \multicolumn{1}{|c|}{\small \textbf{Features}} & \multicolumn{1}{|c|}{\small \textbf{Models}} & \multicolumn{1}{|c|}{\small \textbf{Pros}} & \multicolumn{1}{|c|}{\small \textbf{Cons}} & \multicolumn{1}{|c|}{\small \textbf{Ref.}}\\ 
\hline 
\endfirsthead
\multicolumn{8}{c}{{\bfseries Continued from previous page}} \\
\hline 
\multicolumn{1}{|c|}{\small \textbf{Sn}} & \multicolumn{1}{|c|}{\small \textbf{Application}} & \multicolumn{1}{|c|}{\small \textbf{Year}} & \multicolumn{1}{|c|}{\small \textbf{Features}} & \multicolumn{1}{|c|}{\small \textbf{Models}} & \multicolumn{1}{|c|}{\small \textbf{Pros}} & \multicolumn{1}{|c|}{\small \textbf{Cons}} & \multicolumn{1}{|c|}{\small \textbf{Ref.}}\\ 
\hline 
\endhead
\hline
\endfoot
\bottomrule
\endlastfoot

\multicolumn{8}{c}{\textbf{Pre-harvesting}} \\
\hline
1 & Soil Classification & 1995 & The author is trying to give a quantitative soil classification. & 
$\bullet$ Nueral Network 
& $\bullet$ The effectiveness of this technique is due to its ability to classify soil quantitatively. 
& $\bullet$ Requires a large amount of high-quality training data to achieve accurate results 
& \cite{cal1995soil} \\
\hline
2 & Weather prediction & 2009 & This paper describes the use of SVM for predicting weather. & 
$\bullet$ SVM 
& $\bullet$ The finding that SVM consistently outperformed MLP suggests that SVM may be a more effective approach for weather prediction tasks. 
& $\bullet$ The selection of appropriate parameters is critical for the successful implementation of SVM in weather prediction. 
& \cite{radhika2009atmospheric} \\
\hline
3 & Soil Classification & 2016 & This paper explains SVM based classification of the soil types. & 
$\bullet$ SVM 
& $\bullet$ Model is showing less accuracy 
& $\bullet$ SVM is a powerful ML technique that can handle large and complex datasets.\newline$\bullet$ SVM models are sensitive to the choice of kernel function and its associated parameters, which requires careful tuning to optimize performance. 
& \cite{srunitha2016performance} \\
\hline
4 & Soil Temperature & 2016 & Soil in this paper the measurement is for soil temperature (ST) by AI is presented. & 
$\bullet$ ELM\newline$\bullet$ SaE-ELM\newline$\bullet$ genetic p rogramming (GP)\newline$\bullet$ ANN 
& $\bullet$ Higher generalization performance and prediction accuracy compared to traditional algorithms. 
& $\bullet$ SVM-based soil classification shows lower accuracy compared to alternative classification methods. 
& \cite{nahvi2016using} \\
\hline
5 & Soyabeen pods & 2018 & 38 types of variety of soyabeen pods has been classified. & 
$\bullet$ CNN\newline$\bullet$ SVM 
& $\bullet$ CNNs offer a more accurate and efficient approach for feature extraction compared to traditional methods. 
& $\bullet$ Training and fine-tuning CNNs can be computationally expensive, especially for large datasets. 
& \cite{uzal2018seed} \\
\hline
6 & Seed test & 2019 & This paper proposes a method for classifying plant seedlings using a dataset consisting of around 5,000 images representing 960 unique plants from 12 species at different developmental stages. & 
$\bullet$ CNN 
& $\bullet$ The proposed solution can assist farmers in optimizing their crop production, which can potentially lead to higher yields and profits. 
& $\bullet$ Both CNN and SVM models require careful selection of hyperparameters, which is a time-consuming process and requires expert knowledge. 
& \cite{ashqar2019plant} \\
\hline
7 & Irrigation plaining & 2019 & The proposed solution will be developed by establishing a distributed wireless sensor network, wherein each region of the farm would be covered by various sensor modules which will be transmitting data on a common server. & 
$\bullet$ IoT\newline$\bullet$ SVR\newline$\bullet$ RF\newline$\bullet$ Wireless sendors 
& $\bullet$ Sutaiable and computational efficient approach by Iot and ML 
& $\bullet$ Reliability concerns due to the possibility of sensor failures or connectivity issues. 
& \cite{vij2020iot} \\
\hline
8 & Crop prediction and weed detection & 2020 & The aim of the paper is to provide farmers with a user-friendly and straightforward application that can assist them in selecting the appropriate crops for their land and detecting and eliminating weeds in their crop fields. & 
$\bullet$ CNN\newline$\bullet$ WSN\newline$\bullet$ naive bayes 
& $\bullet$ Precise management of the farm sector can lead to increased productivity and cost savings for farmers. 
& $\bullet$ The model requires significant initial investment and ongoing maintenance costs for hardware, software, and data collection. 
& \cite{dasgupta2020ai} \\
\hline
9 & Seed test & 2020 & A mobile application has been developed that quickly detects and classifies seed images with high accuracy using CNN, one of the deep learning techniques. & 
$\bullet$ CNN\newline$\bullet$ Inceptionv3\newline$\bullet$ Xception\newline$\bullet$ Inception ResNetV2 
& $\bullet$ The mobile application can detect and classify seed images with high accuracy using CNN, a powerful deep learning technique.\newline$\bullet$ The use of advanced models such as Inceptionv3, Xception, and InceptionResNetV2 can enhance the performance of the application. 
& $\bullet$ The system is not able to identify all seed species accurately, especially rare or less common ones. 
& \cite{gulzar2020convolution} \\
\hline
10 & Irrigation plaining & 2020 & The authur has propsed a irrigation system using ML. & 
$\bullet$ CNN\newline$\bullet$ DT 
& $\bullet$ The use of CNN architecture helps to capture complex patterns and relationships in the data.\newline$\bullet$ The decision tree algorithm helps in improving the interpretability and explainability of the model. 
& $\bullet$ The model's performance may be affected by the quality and accuracy of the data used for training, as well as changes in weather conditions. 
& \cite{chen2020deep} \\
\hline
11 & Irrigation planing & 2020 & This paper presents an advanced technology based smart system to predict the irrigation requirements of a field by sensing of ground parameter like moisture of the soil, temperature-humidity and water level using ML algorithm. & 
$\bullet$ RF 
& $\bullet$ The system offers a solution to the over-fitting problem faced by existing algorithms like K-mean and SVM.\newline$\bullet$ The system has the capacity to realize a fully automated irrigation scheme. 
& $\bullet$ The system may not perform optimally under extreme weather conditions or in areas with highly variable soil types. 
& \cite{akshay2020efficient} \\
\hline
12 & Irrigation planing & 2020 & This research presents a smart irrigation system that utilizes ML algorithms to predict the water requirements of a crop. & 
$\bullet$ DT 
& $\bullet$ Better yield prediction 
& $\bullet$ Decision trees are prone to over-fitting the training data. 
& \cite{reddy2020iot} \\
\hline
13 & Soil & 2020 & The study aims to develop a model that can predict soil fertility indices for different villages based on soil reaction, organic carbon, and nutrient content. & 
$\bullet$ ELM with different activation functions like sine-squared, Gaussian radial basis etc. 
& $\bullet$ The approach is cost-effective and efficient, as it eliminates the need for expensive and time-consuming laboratory tests. 
& $\bullet$ Not able to capture the complex interactions between different soil parameters and their impact on soil fertility indices. 
& \cite{suchithra2020improving} \\
\hline
14 & Crop recommendation system & 2021 & The paper presents a yield prediction system for farmers that is both practical and user-friendly. & 
$\bullet$ SVM\newline$\bullet$ ANN\newline$\bullet$ RF\newline$\bullet$ multi linear regression 
& $\bullet$ These models handles large amount of data and complexity of data 
& $\bullet$ The effectiveness of crop selection assistance may depend on the availability and accuracy of data on soil and climate conditions. 
& \cite{9418351} \\
\hline
15 & Seed test & 2021 & This paper shows that wheat seed classification. & 
$\bullet$ Neural network 
& $\bullet$ The model has High accuracy. 
& $\bullet$ Neural networks can be complex to design and train, requiring specialized knowledge and resources. 
& \cite{madhavan2021wheat} \\
\hline
16 & Crop recomendation system & 2022 & The author has presented an analysis of AI-driven precision farming and agriculture, including related studies, and has proposed a new crop recommendation platform that utilizes cloud-based ML. & 
$\bullet$ KNN\newline$\bullet$ DT\newline$\bullet$ RF\newline$\bullet$ XGBoost\newline$\bullet$ SVM 
& $\bullet$ The platform can potentially reduce waste by providing farmers with more accurate information about when to harvest their crops, resulting in fewer spoiled or unsellable crops 
& $\bullet$ The platform requires a significant amount of data to be effective, which may be difficult to obtain in some regions or for certain types of crops. 
& \cite{thilakarathne2022cloud} \\
\hline
17 & Crop recomendation system & 2022 & The proposed technique combines IoT and AI to develop a smart agricultural system. & 
$\bullet$ IoT\newline$\bullet$ layer Perceptron neural Multinetwork 
& $\bullet$ The system can optimize the use of resources such as water and fertilizers, reducing waste and environmental impact. 
& $\bullet$ The system requires regular maintenance and updates to ensure optimal performance, which is time-consuming and expensive. 
& \cite{jagadeeswari2022artificial} \\
\hline
18 & Seed test & 2022 & This research work aims to explore the application of deep learning techniques for seed classification. & 
$\bullet$ MobileNetV2 
& $\bullet$ The choice of using the MobileNetV2 model was based on its efficient memory usage and straightforward architecture. 
& $\bullet$ Less seed class taken 
& \cite{9711662} \\
\hline
19 & Soil Classification & 2022 & The study aimed to assess the effectiveness of ML modelin classifying potential acid sulfate soils in Jutland, Denmark, and also to apply a model-agnostic interpretation method on the resulting CNN model for improved understanding and insights. & 
$\bullet$ CNN 
& $\bullet$ The use of CNNs for the classification of potential acid sulfate soils is a promising approach for contextual mapping in wetland areas. 
& $\bullet$ The CNN model's accuracy and interpretability may be influenced by the quality and quantity of training data used. 
& \cite{beucher2022interpretation} \\
\hline
20 & Crop recomendation system & 2023 & The proposed crop recommendation system, UACR, allows farmers to input their farm location and automatically receive recommendations for the best crops to grow based on various factors such as rain level and soil type. & 
$\bullet$ Iot 
& $\bullet$ The UACR system simplifies the crop recommendation process for farmers, as they only need to input their location. 
& $\bullet$ The absence of edge/fog nodes in some architectures may pose challenges for processing the large amounts of data. 
& \cite{muhammed2022user} \\
\hline
21 & Irrigation plaining & 2023 & The research proposes an IoT-based intelligent control system for smart agriculture that utilizes sensors to collect environmental data and includes an automated irrigation system. & 
$\bullet$ KNN\newline$\bullet$ SVM\newline$\bullet$ DT\newline$\bullet$ RF 
& $\bullet$ The system can provide real-time monitoring of environmental conditions 
& $\bullet$ Large computing costs. 
& \cite{manikandan2023deep} \\
\hline
22 & Intrusion detection system & 2023 & The paper introduces a novel system for detecting intrusions in IoT networks utilized in agriculture, using deep learning models and image-based features. & 
$\bullet$ CNN\newline$\bullet$ PSO for optimization\newline$\bullet$ VGG16\newline$\bullet$ Exception\newline$\bullet$ Inception model 
& $\bullet$ Evaluating the proposed method with different models can help identify potential weaknesses and areas for improvement. 
& $\bullet$ Time-consuming 
& \cite{el2023optimized} \\
\hline
23 & Soil Classification & 2023 & This study demonstrates the application of ML for soil classification using ML. & 
$\bullet$ XGBoost\newline$\bullet$ KNN imputer 
& $\bullet$ Model gives more accurate result 
& $\bullet$ Model is limited to data quality 
& \cite{aydin2023use} \\
\hline
24 & Pepper seeds &  & 15 features classified based on width, length, and projected area, seed weight and density using ML. & 
$\bullet$ MLP\newline$\bullet$ BLR\newline$\bullet$ single feature models 
& $\bullet$ Model has high accuracy in classification 
& $\bullet$ Less accuracy on more features 
& \cite{tu2018selection} \\
\hline
\multicolumn{8}{c}{\textbf{Harvesting}} \\
\hline
25 & Mobile autonomous robot for agricultural operations & 2015 & This technology enables accurate analysis and classification of fruit images, including those with overlapping fruits, using semantic segmentation for individual fruits and instance segmentation for overlapping fruits. & 
$\bullet$ R-CNN 
& $\bullet$ Models saves time of labour 
& $\bullet$ High cost of robots 
& \cite{durmucs2015design} \\
\hline
26 & Fruit Classification & 2018 & We propose an efficient framework for fruit classification using deep learning. & 
$\bullet$ VGG-16 fine-tuned model\newline$\bullet$ CNN 
& $\bullet$ The system can potentially reduce food waste by accurately identifying and classifying fruits 
& $\bullet$ The initial cost of implementing the system is high 
& \cite{parashar2022fruits} \\
\hline
27 & A fully autonomous picking robot & 2018 & The paper presented AI robot for fruits. & 
$\bullet$ AI Perception 
& $\bullet$ Alternative for costly human picker with no damage 
& $\bullet$ Ensure robustness and to identify its shortcomings 
& \cite{de2018development} \\
\hline
28 & Classifying tomato ripeness in the design of a tomato harvesting robot & 2018 & Improved deep learning-based classification method that improves the accuracy and scalability of tomato ripeness with a small amount of training data. & 
$\bullet$ CNN\newline$\bullet$ DNN 
& $\bullet$ Less parameter calculation and higher accuracy 
& $\bullet$ Over-fitting problem 
& \cite{zhang2018deep} \\
\hline
29 & Automated fruit harvesting robot & 2019 & This research introduces an approach for the identification and harvesting of fruits using a robotic arm. & 
$\bullet$ CNN\newline$\bullet$ R-CNN\newline$\bullet$ VGG\newline$\bullet$ YOLO 
& $\bullet$ The algorithm for fruit position detection and harvesting can be automated 
& $\bullet$ High cost of model 
& \cite{onishi2019automated} \\
\hline
30 & RTFD real time fruit detection & 2019 & This study employs novel convolutional deep learning techniques based on single-shot detectors for fruit (apple and pear) detection and counting within the tree canopy. & 
$\bullet$ CNN\newline$\bullet$ YOLO 
& $\bullet$ Redues fruit losses 
& $\bullet$ The limitation of the model in differentiating between two objects of the same category in the same grid cell may result in inaccurate fruit counting in some cases 
& \cite{bresilla2019single} \\
\hline
31 & Method for mango detection and counting & 2019 & This study addresses the challenge of stable fruit detection in varying illumination conditions present in open farm environments. & 
$\bullet$ CNN 
& $\bullet$ MangoNet, a deep CNN architecture using semantic segmentation Stable fruit detection in varying illumination conditions can improve efficiency and accuracy of fruit counting and yield estimation in open farm environments 
& $\bullet$ This work only focused on one fruit, so the model will not perform well for other fruits. 
& \cite{kestur2019mangonet} \\
\hline
32 & Fruit feature detection & 2019 & Use of (DCNN) to optimize features for classification and segmentation tasks in agricultural navigation. & 
$\bullet$ VGG16 (OxfordNet)\newline$\bullet$ Dcnn 
& $\bullet$ The use of DCNNs and VGG16 for feature optimization and classification improve accuracy in agricultural navigation tasks. 
& $\bullet$ Models work for specific crops only 
& \cite{bakken2019end} \\
\hline
33 & Harvesting sweet pepper fruit in greenhouses & 2019 & The framework presented in this study incorporates three classification models that enable real-time classification of date fruit images based on their type, maturity, and decision regarding harvesting. & 
$\bullet$ AlexNet and VGG-16\newline$\bullet$ CNN 
& $\bullet$ The model achieves best accuracy in fruits classifying 
& $\bullet$ The fine-tuning process of the VGG-16 model requires significant computational resources and expertise to achieve optimal results. 
& \cite{altaheri2019date} \\
\hline
34 & Fruit Yield & 2019 & This paper presents the deep multi-faced system to increase banana fruits yield. & 
$\bullet$ LSTM\newline$\bullet$ DL 
& $\bullet$ The model achieved better performance 
& $\bullet$ Acquiring and curating such data can be time-consuming and resource-intensive 
& \cite{rebortera2019enhanced} \\
\hline
35 & Fruit Detection & 2020 & FD should be robust to varying outdoor conditions with high signal to noise ratio. FD should help in estimating the angle at which the fruit is oriented around the stem for effective severing. & 
$\bullet$ Semantic Segmentation using FCN\newline$\bullet$ Adaptive Image thresholding using video images using reinforced learning with decaying e-greedy algorithm 
& $\bullet$ FD system's robustness and ability to handle varying outdoor conditions contribute to accurate fruit detection 
& $\bullet$ Computational complexity 
& \cite{arad2020development} \\
\hline
36 & Fruit picking robots, & 2020 & The system possesses the ability to train with multiple fruit types, enabling accurate classification based on factors such as size and ripeness. & 
$\bullet$ Improved efficiency 
& $\bullet$ Multiple fruit crops support 
& $\bullet$ Efficiency can be improved 
& \cite{bogue2020fruit} \\
\hline
37 & Fruit picking & 2020 & This article presents a dual-arm aubergine harvesting robot consisting of two robotic arms configured in an anthropomorphic manner to optimize the dual workspace. & 
$\bullet$ Cubic SVM\newline$\bullet$ water shed transformation\newline$\bullet$ Point cloud extraction 
& $\bullet$ The use of a dual-arm configuration in the aubergine harvesting robot allows for optimized workspace 
& $\bullet$ Computational demands may limit real-time performance, requiring specialized hardware 
& \cite{sepulveda2020robotic} \\
\hline
38 & Fruit Detectetion & 2020 & The system demonstrates effectiveness in detecting fruits even in environments where leaves or branches have similar colors, accommodating variability in fruit shape and irregularities in the growing environment. & 
$\bullet$ CNN\newline$\bullet$ SVM 
& $\bullet$ Improved cucumber harvesting efficiency through effective fruit detection and high recognition accuracy 
& $\bullet$ Limited adaptability for diverse agriculture applications 
& \cite{mao2020automatic} \\
\hline
39 & Fruit Detectetion & 2020 & To locate \& harvest cotton balls in field conditions. & 
$\bullet$ Deep learning with Tiny Yolov3 with 7 conv. layers\newline$\bullet$ CLoDSA based image augmentation 
& $\bullet$ Over $77\%$ Action Success Ratio (ASR) and over $97\%$ Detection Performance 
& $\bullet$ Future design and development research should also include alternative energy sources to decrease energy costs 
& \cite{fue2020extensive} \\
\hline
40 & Fruit picking & 2020 & The paper introduces a harvesting robot designed specifically for plucking tea leaves, utilizing Jaco 2 robotic arm. & 
$\bullet$ Jaco 2 based leaf plucking using ProMP 
& $\bullet$ Increased efficiency and reduced labor in tea production. 
& $\bullet$ Found limitations in handling branches with varying stiffness 
& \cite{motokura2020plucking} \\
\hline
41 & Autonomous path detection & 2020 & The paper presents a system for autonomous path detection in semi-structured environments to enhance navigation and operation efficiency. & 
$\bullet$ CNN 
& $\bullet$ Potential for cost savings by minimizing human labor and errors. 
& $\bullet$ Poor lighting, occlusions, and complex surroundings can limit the accuracy of autonomous path detection 
& \cite{kim2020path} \\
\hline
42 & Fruit Detection & 2020 & Improved detection of small harvesting cups using gray scale-depth and color images, with practical lighting-independent applications. & 
$\bullet$ Faster RCNN with MobileNetV2 
& $\bullet$ Accurate detection of small harvesting cups in a wide range of lighting conditions, enabling more efficient and effective operations. 
& $\bullet$ Dependency on color images for tapping position detection 
& \cite{wongtanawijit2019rubber} \\
\hline
43 & Fruit Detection & 2023 & In this research, we propose a lightweight fruit detection algorithm for edge CPUs, enabling 19 FPS detection speed in RTFD-CPU smartphone app. & 
$\bullet$ DNN\newline$\bullet$ PicoDet-S model 
& $\bullet$ The proposed RTFD has great potential for intelligent picking machines, 
& $\bullet$ Sensitivity to environmental factors that may affect fruit detection accuracy 
& \cite{mao2023real} \\
\hline
\multicolumn{8}{c}{\textbf{Post-harvesting}} \\
\hline
44 & Better post harvest handling & 2020 & Improving harvest quality and productivity for farmers through innovative solutions. & 
$\bullet$ Computer vision DL\newline$\bullet$ YOLOV3 
& $\bullet$ Assisting in the identification of healthy apples by distinguishing them from those with defects 
& $\bullet$ Limitation in training different type of data 
& \cite{valdez2020apple} \\
\hline
45 & Better post harvest handling & 2020 & A ML approach utilizes radar back scattering data to accurately estimate grain quantity in silos. & 
$\bullet$ KNN 
& $\bullet$ Easy to apply in different grains 
& $\bullet$ Challenges in accuaring correct radar data 
& \cite{duysak2020machine} \\
\hline
46 & Food security & 2021 & A hierarchical model is proposed, integrating ontology-enabled IoT, to distinguish and separate healthy Sekai-ichi apples using Boosted Continuous Non-spatial Whole Attribute Extraction (BCNAE). & 
$\bullet$ IoT\newline$\bullet$ ML 
& $\bullet$ Achieving higher accuracy and throughput while minimizing processing time 
& $\bullet$ Computational complexity and accuracy-speed trade-off are potential drawbacks when aiming for higher accuracy, throughput, and speed 
& \cite{sanjeevi2021ontology} \\
\hline
47 & Better post harvest handling & 2022 & The review explores the use of sensors, IoT principles, and ML technologies in post-harvest grain monitoring, identifying potential advancements and resource utilization. & 
$\bullet$ IoT 
& $\bullet$ Enhanced grain quality monitoring 
& $\bullet$ Collecting sensor data requires robust measures to protect sensitive information. 
& \cite{lutz2022applications} \\
\hline
48 & Food security & 2022 & The paper has investigated household-level food security. & 
$\bullet$ RF\newline$\bullet$ decision tree 
& $\bullet$ Prone to over-fitting 
& $\bullet$ The effectiveness of the model vary across different locations 
& \cite{ahn2022food} \\
\hline
49 & Food security & 2022 & The paper has proposed detection and prediction models for identifying post-harvest deterioration zones in stored apple fruits. & 
$\bullet$ CNN-based U-Net\newline$\bullet$ Deep-Lab\newline$\bullet$ Mask R-CNN 
& $\bullet$ The models enable retailers to maintain better quality control of their apple fruits, ensuring customer satisfaction. 
& $\bullet$ The models are specifically designed for stored apple fruits only 
& \cite{singh2022recent} \\
\hline
50 & Food security & 2022 & The paper has presented Drone technology for food supply chain management. & 
$\bullet$ DL\newline$\bullet$ ML 
& $\bullet$ Reduces costs associated with traditional packing and delivery methods 
& $\bullet$ Drones have limited payload capacity, battery life issues 
& \cite{kler2022machine} \\
\hline
\end{longtable}

            }
    \end{landscape}

Table \ref{breeding} presents an overview of recent applications of ML techniques in the realm of agricultural practices, with a specific emphasis on genetic changes and plant breeding. The table chronicles research spanning from 2018 to 2023 and encapsulates various ML models. These models have been used to accelerate crop improvement, aid in precision decision making for rapid breeding, improve climate-resilient crop breeding, and even for automated plant phenotyping. Several references, such as \cite{harfouche2019accelerating} and \cite{yoosefzadeh2023machine}, further emphasize the significance of ML in advancing modern agricultural practices and outcomes.

  \begin{table*}[!htp]\centering
    \caption{Overview of Recent Machine Learning Applications in Agricultural Practices, including Genetic Changes and Plant Breeding.}
    \label{breeding}
    \scriptsize
    \renewcommand{\arraystretch}{1.3} 
    \begin{tabular}{|p{2.0cm}|p{0.5cm}|p{3cm}|p{6.5cm}|p{0.5cm}|}
    \toprule
    \midrule
    \textbf{Application} & \textbf{Year} & \textbf{Model} & \textbf{Features} & \textbf{Ref.} \\
    \hline
    Plant Breeding & 2019 & 
    $\bullet$ Neural Network \newline
    $\bullet$ Explainable AI 
    & The author has presented the state of genomics and field phenomics and proposes a workable path to improvement. & \cite{harfouche2019accelerating} \\
    \hline
    Plant Breeding & 2023 & 
    $\bullet$ Ensemble Learning \newline
    $\bullet$ Big Data \newline
    $\bullet$ Random Forest  
    & ML has been used in analyzing genetic data in plant breeding, accelerating crop improvement, developing new varieties, and ensuring food data security. & \cite{yoosefzadeh2023machine} \\
    \hline
    Integrating Speed Breeding & 2022 & 
    $\bullet$ Deep Learning \newline
    $\bullet$ Neural Network  
    & The author has utilized ML in speeding plant breeding and enabling precision decision-making. & \cite{rai2022integrating} \\
    \hline
    Climate-Resilient Smart-Crop Breeding & 2022 & 
    $\bullet$ Machine Learning  
    & The author has reviewed ML applications in crop breeding. & \cite{khan2022applications} \\
    \hline
    Plant Breeding & 2020 & 
    $\bullet$ Neural Network \newline
    $\bullet$ SVM \newline
    $\bullet$ Random Forest  
    & Discussed improving new layers of plant breeding. & \cite{niazian2020machine} \\
    \hline
    Plant Breeding & 2021 & 
    $\bullet$ Multilayer Perceptron \newline
    $\bullet$ SVM \newline
    $\bullet$ Random Forest  
    & ML models have been used for improving soybean crop yield. & \cite{yoosefzadeh2021application} \\
    \hline
    Plant Breeding & 2019 & 
    $\bullet$ Random Forest  
    & Predicted in-season seed yield (SY) in soybean based on high-dimensional phenotypic trait data. & \cite{parmley2019machine} \\
    \hline
    Plant Breeding & 2019 & 
    $\bullet$ SVM \newline
    $\bullet$ Deep Learning  
    & Compared the performance of MLP and SVM methods to the Bayesian threshold genomic best linear unbiased prediction (TGBLUP) model for genomic selection in plant breeding. & \cite{montesinos2019benchmarking} \\
    \hline
    Estimating Phenotypic Traits & 2018 & 
    $\bullet$ Regression \newline
    $\bullet$ Deep Learning  
    & Using color images and elevation maps, ML is employed to estimate emergence and biomass traits in wheat field plots. & \cite{8354146} \\
    \hline
    Phenotyping & 2019 & 
    $\bullet$ Deep Learning \newline
    $\bullet$ CNN  
    & CNN-based modeling is used in computer vision for automated plant phenotyping, enabling high-throughput analysis of plant characteristics. & \cite{mochida2019computer} \\
    \midrule
    \bottomrule
    \end{tabular}
\end{table*}

\subsection{Machine Learning  Application in Disease Detection}
Researchers have extensively used ML and image processing techniques to identify diseases in various types of plants during pre-harvest, harvest, and post-harvesting. Hence, a separate discussion is required to address this focus.

In \cite{das2020leaf} an automated approach is proposed for the detection of leaf disease in agriculture for tomato plants. The model aims to detect diseases in early stages, enhancing crop production. SVM, Random Forest, and Logistic Regression algorithms are compared, and SVM outperforms the others. The study lacks specific details on the size of the data set and the evaluation metrics. Similarly, a system for the detection of apple disease based on ML and computer vision is proposed in \cite{poojary2018survey}. It achieves $98.63\%$ accuracy by segmenting diseased regions and utilizing features such as DWT and color histograms.

For accurate and efficient multi-class plant disease detection, a deep learning object detection model is proposed in \cite{weizheng2008grading}, surpassing the performance of existing models with a mean average precision of $91.2\%$ and an F1-score of $95.9\%$. Another study \cite{jiang2019real} introduces a deep convolutional neural network (DCNN) for early diagnosis of apple tree leaf diseases (ATLDs). The proposed DCNN achieves $98.82\%$ accuracy, showing fast convergence, fewer parameters, and high robustness.
In \cite{hasan2022disease}, an accurate identification approach for common apple leaf diseases is proposed using a deep convolutional neural network. It achieves a $97.62\%$ accuracy with reduced parameters. A lightweight convolutional neural network, RegNet, is introduced in \cite{li2022apple} for rapid and accurate identification of apple leaf diseases, surpassing other pre-trained models with a $99.23\%$ overall accuracy.
The research in \cite{chao2020identification} presents a CNN model with $19$ layers for accurate classification of Marsonina Coronaria and Apple Scab diseases in apple leaves, achieving $99.2\%$ accuracy on a dataset of $50,000$ images. Another ML-based approach for the automated detection of apple tree diseases is proposed in \cite{liu2017identification}, achieving a high accuracy of approximately $95\%$ for the classification of diseases. 
A deep convolutional neural network model is proposed in \cite{aich2023automated} for real-time detection of apple leaf diseases, achieving an overall accuracy of $99.31\%$ and surpassing other models. The research in \cite{mahato2022improved} utilizes fine-tuned pre-trained CNN models for accurate apple crop disease detection, with DenseNet201 achieving $98.75\%$ accuracy.

In \cite{pradhan2022comparison}, an automated software solution is proposed using image processing techniques for the detection of apple leaf disease, based on color and texture characteristics. Finally, an algorithm for accurate segmentation of diseased parts in apple leaf images is proposed in \cite{singh2022extraction}, achieving an accuracy of $96.4\%$ with the $k$ nearest-neighbor classifier. A detailed look is given in Table~\ref{tab:disease_detection}.

\begin{table*}[ht]
    \centering
    \caption{Disease Detection Applications in Agriculture (Pre-harvesting and Post-harvesting)}
    \label{tab:disease_detection}
    \scriptsize
    \renewcommand{\arraystretch}{1.3} 
    \begin{tabular}{|p{0.5cm}|p{3cm}|p{0.7cm}|p{5.0cm}|p{2.0cm}|p{0.5cm}|}
    \toprule
    \midrule
    \textbf{Sn} & \textbf{Application} & \textbf{Year} & \textbf{Features} & \textbf{Models} & \textbf{Ref.} \\
    \hline
    
    \hline
        1 & Pre-harvesting disease detection & 2020 & CNN is used to automatically identify crop diseases & 
        $\bullet$ Inception-ResNet-v2 
        & \cite{ai2020research} \\
        \hline
        2 & Pre-harvesting disease detection and crop recommendation system & 2021 & Presents a location and season-based crop recommendation system. & 
        $\bullet$ CNN \newline
        $\bullet$ VGG-16 
        & \cite{patil2021krushi} \\
        \hline
        3 & Pre-harvesting disease detection & 2021 & A hybrid system based on a Convolutional Autoencoder (CAE) network and CNN for automatic plant disease detection. & 
        $\bullet$ CNN 
        & \cite{bedi2021plant} \\
        \hline
        4 & Pre-harvesting disease detection & 2021 & Uses computer vision and AI for early detection of plant diseases, reducing continuous human monitoring challenges. & 
        $\bullet$ CNN 
        & \cite{chowdhury2021automatic} \\
        \hline
        5 & Pre-harvesting disease detection & 2022 & A model using CNN and VGG to accurately classify disease-affected leaves. & 
        $\bullet$ CNN \newline
        $\bullet$ VGG 
        & \cite{paymode2022transfer} \\
        \hline
        6 & Pre-harvesting disease detection & 2022 & Employs a hybrid ML model and a web-based AI tool for plant disease identification. & 
        $\bullet$ CNN \newline
        $\bullet$ AlexNet \newline
        $\bullet$ LeNet 
        & \cite{mridha2021plant} \\
        \hline
        7 & Post harvesting disease detection & 2011 & Automatically extracts quality features from external data using image processing. & 
        $\bullet$ CNN 
        & \cite{al2011computer} \\
        \hline
        8 & Post harvesting disease detection & 2017 & Incorporates a prediction system for warehouse managers to recommend optimal dispatch sequences and transportation conditions. & 
        $\bullet$ IoT \newline
        $\bullet$ ML 
        & \cite{purandare2016analysis} \\
        \hline
        9 & Post harvesting disease detection & 2019 & Uses image processing techniques to classify mangoes based on their condition across three phases. & 
        $\bullet$ ML \newline
        $\bullet$ DOF \newline
        $\bullet$ Edge Detection \newline
        $\bullet$ RGB 
        & \cite{chakravarthy2019micro} \\
        \hline
        10 & Post harvesting disease detection & 2019 & Detects calyx and stalk scars in both defected and healthy tomatoes. & 
        $\bullet$ RBF SVM 
        & \cite{ireri2019computer} \\
        \hline
        11 & Post harvesting disease detection & 2023 & Introduces a knowledge-based expert system utilizing a Bayesian network for diagnosing post-harvest diseases in apples. & 
        $\bullet$ Bayesian network 
        & \cite{sottocornola2023development} \\
    \hline
    \bottomrule
    \end{tabular}
\end{table*}

\subsection{Model used in Literature }

To provide deeper insights into the utilization of ML models in literature, here is a detailed analysis of models applied across various domains from Table \ref{tab:long} and  \ref{breeding}. 

Neural networks (NN) and SVM are widely applied for soil classification \cite{aydin2023use} and weather prediction \cite{radhika2009atmospheric}, offering significant accuracy and efficiency. However, their success depends heavily on parameter tuning and data quality. ML models integrated with IoT and wireless sensor networks have optimized irrigation planning, offering real-time monitoring and efficient water usage. Similarly, crop recommendation systems employ ML models like RF, SVM, and deep learning to provide farmers with practical decision-making tools, considering soil type and environmental factors \cite{van2020crop}. However, issues such as sensor reliability and maintenance costs remain challenges.

CNNs and computer vision have been extensively used for detecting plant diseases and classifying seeds with high accuracy \cite{paymode2022transfer}. Advanced architectures like Inception-ResNetV2 and MobileNetV2 have further enhanced performance \cite{ai2020research, 9711662}. In the domain of fruit detection and harvesting, models like YOLO and R-CNN have been pivotal in automating the tasks of detection in image data, reducing labor dependency, and improving efficiency \cite{fue2020extensive, onishi2019automated}.  LSTM models have been mostly used for analyzing and predicting crop yield \cite{rebortera2019enhanced}. 

ML models have accelerated crop improvement by analyzing genetic data, predicting seed yield, and enabling precision breeding. Random Forest and Neural Networks, in particular, have proven effective in estimating phenotypic traits and improving crop yield \cite{yoosefzadeh2023machine}. Additionally, explainable AI is emerging in plant breeding to enhance model interpretability and decision-making \cite{harfouche2019accelerating}. 

The analysis highlights that CNNs dominate in image-based applications such as disease detection and fruit classification, while IoT-integrated models are prevalent in irrigation and post-harvest management. Traditional models like SVM and Random Forest are widely used due to their simplicity and efficiency, particularly for soil and weather-related applications.  These findings emphasize the importance of selecting appropriate ML models for specific agricultural challenges while addressing their limitations to maximize their practical applicability and impact on sustainable farming practices.

    \begin{figure*}[htbp]
        \centering
        \includegraphics[width=\linewidth, height=0.75\textheight, keepaspectratio]{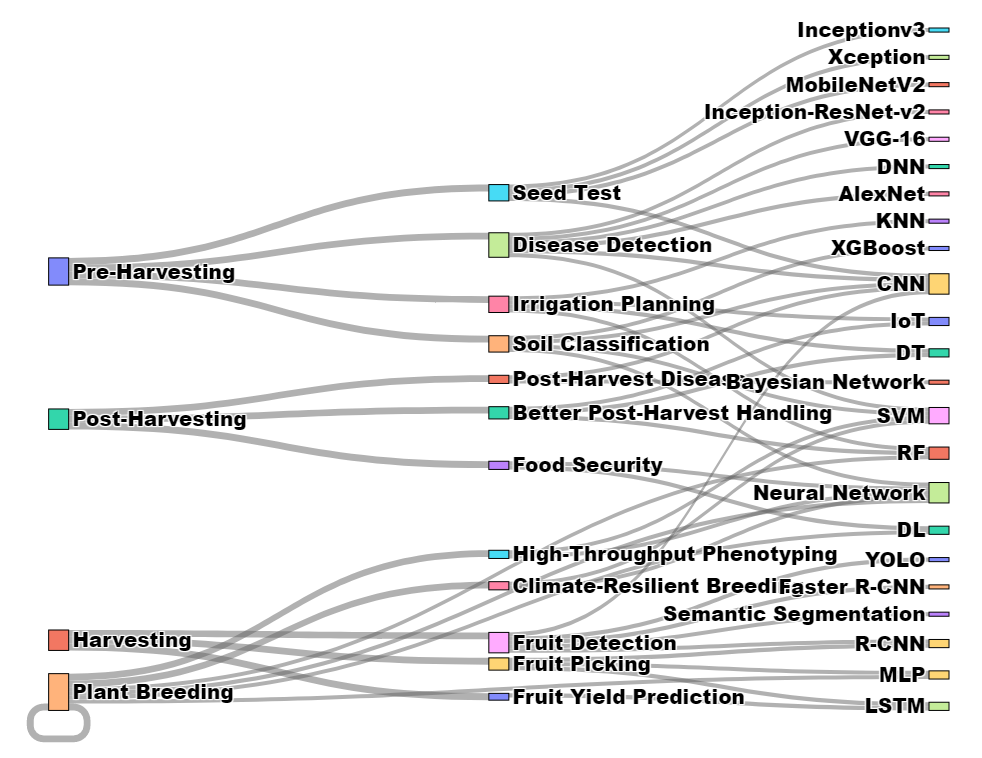}
        \caption{Sankey diagram illustrating the integration of various ML models and technologies (e.g., IoT, YOLO) across different stages of agricultural processes, including pre-harvest, post-harvest, harvest handling, and plant breeding, contributing to improved food security and sustainable farming practices.}
        \label{Lit_model}
    \end{figure*}

A concise visualizes the application of ML models across agricultural phases, based on insights from the literature (Table~\ref{tab:long} and \ref{breeding}) is given in the sankey diagram Fig.\ref{Lit_model}. From the diagram, it is clear that, in the pre-harvesting phase, ML models like CNN, SVM, and XGBoost are widely used for soil classification, seed testing, irrigation planning, and disease detection. These models support precision farming by analyzing soil properties, optimizing water use, and identifying crop diseases early. Harvesting relies heavily on computer vision models such as YOLO, R-CNN, and LSTM for fruit detection, picking, and yield prediction, ensuring automated and efficient crop assessment.

Post-harvesting integrates IoT, Bayesian Networks, and DL models for disease detection, food security, and handling, focusing on supply chain efficiency and quality control. Plant breeding involves phenotyping and climate-resilient breeding, leveraging Neural Networks, SVM, and DL to enhance crop adaptability and productivity.

\subsubsection{Comparison of Machine Learning Techniques in Agriculture}

ML models are widely used in agriculture for image processing, classification, prediction, and automation. Different ML techniques are suited to specific tasks. This section consolidates ML methods to avoid redundancy in subsections \ref{sec:ml_Pre_Harvesting}, \ref{sec:ml_harvesting},  \ref{sec:ml_Post_harvesting}.

\begin{itemize}
    \item Convolutional Neural Networks (CNNs): CNNs are used for tasks where image data is being used such as crop disease detection, fruit classification, and yield estimation. CNNs provide high accuracy in feature extraction but require large datasets and high computational power \cite{patil2021krushi} .
    \item Support Vector Machines (SVMs): In ML classification tasks, including soil type identification and pest detection. SVMs work well with structured datasets but struggle with complex, high-dimensional data \cite{manikandan2023deep, sepulveda2020robotic}.
    \item K-Nearest Neighbors (K-NN): Used in soil moisture prediction and crop classification due to its ease of implementation. However, it becomes computationally expensive for large datasets  \cite{manikandan2023deep}.
    \item Decision Trees \& Random Forest: These ensemble learninf based algorithms re effective for yield prediction and irrigation optimization. These models handle missing data well and provide interpretable decision-making \cite{mao2020automatic}.
    \item Extreme Learning Machines (ELM): ELMs are mostly applied in soil temperature estimation and regression-based problems due to their fast training speed \cite{peng2024grapeleaf} .
    \item Deep Learning \& Hybrid Models: These are advanced architectures that integrate CNNs, RNNs, and transformers for real-time agricultural monitoring and automation \cite{bayoudh2024survey}.
\end{itemize}

\section{Information Fusion in Agriculture}\label{Sec 4}

To improve the accuracy of agricultural practices, researchers and practitioners integrate data from diverse sources—a process known as information fusion in agriculture \cite{allu2025remotesensing}. Addressing \textbf{RQ-2}, this section examines common dataset sources and the types of datasets used in agricultural applications. It further discusses how ML-driven data fusion enhances agricultural outcomes through ensemble learning, hybrid models, data aggregation strategies, and transfer learning.

\begin{figure}[htbp]
        \centering
        \includegraphics[width=0.99\textwidth]{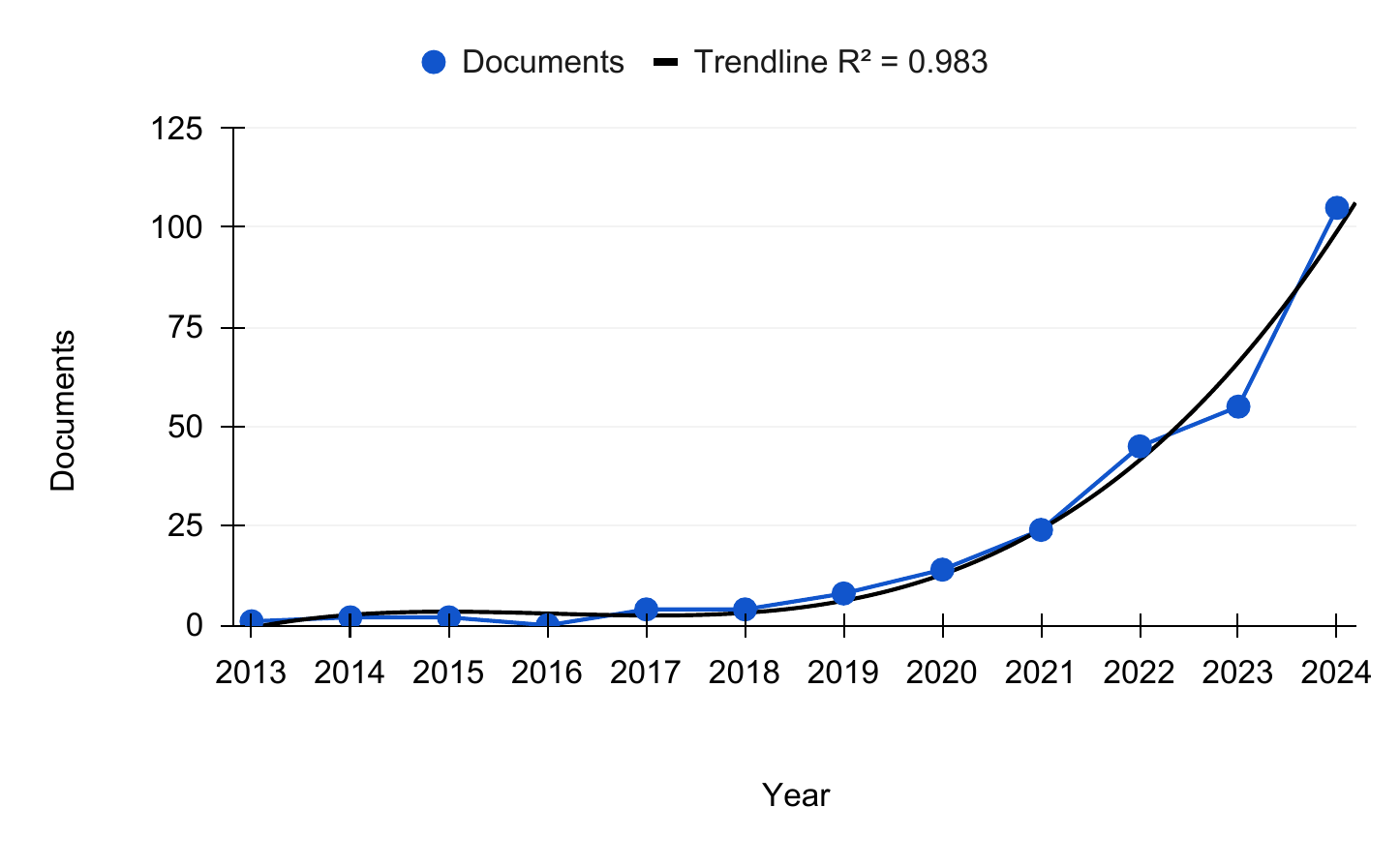}
        \caption{Trends in Multi-Source Data Fusion for Agricultural Machine Learning. Data source: Scopus, 30 December 2024.}
        \label{research_fusion}
    \end{figure}

A database search in Scopus was performed on 30 December 2024 using the keywords ``machine learning", ``data fusion", and ``agriculture" (searched within the title, abstract, or keywords). The resulting trends, depicted in Fig.~\ref{research_fusion}, reveal a steady year-by-year increase in research on data fusion.
    
\subsection{Common Data Types in Agriculture}

The common data types utilized in agriculture with ML, including RGB images, hyperspectral images, multispectral images, and text data, are summarized in Table \ref{tab:data_categories}. For instance, the study by Maimaitijiang et al. \cite{maimaitijiang2020soybeanyield} evaluates the effectiveness of UAV-based multimodal data fusion—integrating RGB, multispectral, and thermal sensors—for estimating soybean (\textit{Glycine max}) grain yield within a Deep Neural Network (DNN) framework. The researchers collected RGB, multispectral, and thermal images using a cost-effective multi-sensor UAV at a test site in Columbia, Missouri, USA.  The study highlights three key findings:  
    \begin{itemize}
        \item Multimodal data fusion enhances yield prediction accuracy and improves adaptability to spatial variations.  
        \item DNN-based models outperform traditional methods, with the DNN-F2 model achieving the highest performance (\( R^2 = 0.720 \) and RMSE\% = 15.9\%).  
        \item DNN models demonstrate greater resilience to saturation effects and exhibit more adaptive performance across different soybean genotypes.
    \end{itemize}

Remote sensing data plays a critical role in precision agriculture, with key sources of data acquisition including satellite imagery, aerial data, drone (UAV) data, and LiDAR \& radar data \cite{allu2025remotesensing} . These data types are widely used for crop monitoring, disease detection, and yield prediction through multispectral, hyperspectral, and thermal imaging. Time-series data, primarily collected from IoT sensors, is crucial for monitoring environmental parameters such as temperature, humidity, and soil moisture  \cite{rahman2023timeseries}. Tabular data, consisting of structured information like soil composition, crop characteristics, and historical weather records, supports predictive analytics for yield estimation and market forecasting \cite{richetti2023deeplearning}. Text data, derived from farmer surveys and agricultural reports, helps in shaping more sustainable policies and providing farmers with accurate feedback, as seen in AI-based agricultural chatbot Kisan GPT \cite{rehman2022semantics}. Additionally, biological data, including genetic and phenotypic information, is essential for plant breeding, predicting plant traits, and estimating phenotypic characteristics for improvement of different crop varieties \cite{harfouche2019accelerating}.

    \begin{table*}[!htp]\centering
        \caption{Common Data Types and Data Modalities Used in Agriculture with  Machine Learning}
        \label{tab:data_categories}
        \scriptsize
        \begin{tabular}{|p{3cm}|p{3cm}|p{5cm}|p{1cm}|}
            \toprule
            \midrule
            \multicolumn{2}{|c|}{\textbf{Data Category and Subcategory}} & \textbf{Data Modality} & \textbf{Ref.} \\
            \hline
            \multirow{5}{=}{Remote Sensing Data} 
            & Satellite Data & Image Data (Multispectral, Hyperspectral, Thermal, Radar) & \cite{allu2025remotesensing} \\
            & Aerial Data & Image Data (RGB, Thermal) & \\
            & Drone (UAV) Data & Image Data (RGB, Multispectral, Hyperspectral, Thermal) &  \\
            & LiDAR \& Radar Data & No Traditional Images (3D Point Clouds \& Reflectance Data) &  \\
            & IoT Sensors with Spectral Imaging & Image Data &  \\
            \hline
            Time-Series Data & IoT Sensors Data & Numerical Data & \cite{rahman2023timeseries} \\
            \hline
            Tabular Data &  & Numerical \& Categorical Data & \cite{richetti2023deeplearning} \\
            \hline
            Text Data &  & Text Data & \cite{rehman2022semantics} \\
            \hline
            \multicolumn{2}{|p{7cm}|}{{Biological Data (Genetic Data, Phenotypic Data)}} 
            & Image Data (RGB, Hyperspectral, Thermal), Tabular \& Numerical Data & \cite{harfouche2019accelerating} \\
            \midrule
            \bottomrule
        \end{tabular}
    \end{table*}

\subsection{Common Machine Learning Techniques for Data Fusion}

\begin{itemize}

    \item\textbf{Ensemble Learning for Multi-Source Data Integration:}  
    Ensemble learning methods, combines multiple models to improve prediction accuracy and robustness. Techniques such as Bagging, Boosting, and Stacking, help handle multi-source data by reducing individual model biases and improving the generalizability of predictions by ML models \cite{yang2024dane}.  In the paper \cite{chen2022appleyield}, the authors have used ensemble learning methods for yield prediction with (light detection and ranging) LiDAR and  Multispectral imagery data from UAV (unmanned aerial vehicles). While in the paper \cite{nguyen2022soilcarbon}, Extreme Gradient Boosting (XGBoost) has been used in monitoring agricultural soil organic carbon (SOC) with Sentinel-1 (S1) C-band dual polarimetric synthetic aperture radar (SAR) and Sentinel-2 (S2) multispectral data.

    \item\textbf{Bayesian Inference for data fusion:} 
    One of the common method for multi sensor data fusion is Bayesian theory that helps make better decisions on bridges. It is also used to image fusion of same scene into a single image that contains all the imporatnt features from each of the original images. The resulting images will be more suitable for machine leanring  models for further decision making \cite{sciencedirect2025bayesian}. In the paper \cite{khan2024tomatoleaf}, the authors have proposed seven robust Bayesian optimized deep hybrid learning models for the automated classification of ten types of tomato leaves with testing of popular Plantvillage datasets.

    \item\textbf{Hybrid Models for Cross-Modal Learning:}  
    When multiple ML or deep learning (DL) architectures are combined to handle data from different sources, the resultant model is called a multimodal hybrid ML model. These methods help extract more features from multi-modal data, and the combined features are known as hybrid features. However, selecting the optimal fusion techniques for creating hybrid models remains critical for achieving optimal model performance \cite{bayoudh2024survey}. The study \cite{dai2023wolfberry} proposed a hybrid model named ITF-WPI for multimodal pest identification and classification in wolfberry. This model utilizes multimodal data (images and text), incorporating Transformer-based contextual extraction (CoT) and LSTM models for improved performance.

    \item\textbf{Data Aggregation Strategies for Large-Scale Agricultural Data:}  
    Data aggregation techniques merge data from multiple sources, such as sensor networks, satellite imagery, and IoT devices, into a unified format for ML models. Aggregating this type of data is crucial, as it becomes large-scale and complex, requiring structured processing for informed decision-making. These strategies include hierarchical aggregation, spatial aggregation and  temporal aggregation , explain below briefly \cite{nabil2022data}. In this paper \cite{yuan2020dataaggregation}, the authors propose an efficient and scalable framework for privacy-preserving data aggregation in smart agriculture, utilizing data from smart devices and IoT networks. However the paper \cite{song2020fpdp}, explains a data aggregation scheme based on the ElGamal Cryptosystem ensuring, secure and private data processing of agricultural data taken with the use of IoT and 5th generation wireless network (5G).

    \item\textbf{Deep Learning-based Feature Fusion and Transfer Learning for Cross-Modal Data Adaptation:}
    Deep learning-based feature fusion involves techniques that automatically extract and combine features from multiple sources, such as images, text, or sensors, using deep neural networks to improve predictive performance and enhance decision-making \cite{sciencedirect2025featurefusion}.
    
    In addition, transfer learning for cross-modal data adaptation allows ML models to leverage knowledge from one data modality (e.g., satellite imagery) and apply it to another (e.g., IoT sensor readings or time-series data). By fine-tuning pre-trained models for new tasks, this approach facilitates cross-domain adaptation and boosts performance in specific applications, such as precision agriculture, environmental monitoring, and smart city infrastructure \cite{zhen2020deep}. For example, in the study \cite{fan2022leaf}, a general framework has been proposed to extract high-level latent feature representations using transfer learning methods.

\end{itemize}

\subsection{Applications of Data Fusion and Hybrid Models in Various Agricultural Domains}
This section highlights successful studies utilizing data fusion and hybrid ML models across various agricultural domains. While advanced fusion techniques are rapidly evolving in agriculture \cite{yashodha2021iot, pantazi2019intelligent}, the literature still lacks extensive research in this area. This section discusses recent studies that incorporate data fusion and hybrid ML/DL models for applications such as crop monitoring, disease detection, yield prediction, crop identification and crop classification, and land monitoring in the Table~\ref{tab:ml_techniques}.


\begin{table*}[!htp]\centering
    \caption{Some Studies on Data Fusion and Hybrid Models in Various Agricultural Domains}
    \label{tab:ml_techniques}
    \scriptsize
    \renewcommand{\arraystretch}{1.3} 
    \begin{tabular}{|p{3.9cm}|p{3cm}|p{2cm}|p{1cm}|p{1.2cm}|p{0.5cm}|} 
        \hline
        \textbf{ML Technique} & \textbf{Data / Sensor Used} & \textbf{Application Areas} & \textbf{Data Fusion} & \textbf{Hybrid Models} & \textbf{Ref.} \\
        \hline \hline
        $\bullet$ Partial Least Squares (PLS) Regression-Based Feature Fusion \newline
        $\bullet$ Ensemble Learning Model
        & PlantVillage Data 
        & Crop Disease Recognition 
        & $\times$  
        & \checkmark  
        & \cite{saeed2021cropdiseases} \\
        \hline
        $\bullet$ Deep Neural Networks
        & RGB, Multispectral, and Thermal Sensor Data 
        & Soybean Yield Prediction Using Low-Cost Sensors
        & \checkmark  
        & $\times$ 
        & \cite{maimaitijiang2020soybeanyield} \\
        \hline
        $\bullet$ Extreme Learning Machine (ELM)
        & Multispectral Data 
        & Grape Leaf Moisture Prediction 
        & \checkmark  
        & \checkmark  
        & \cite{peng2024grapeleaf} \\
        \hline
        $\bullet$ Transformer-Based GANet Model
        & RGB, Thermal Infrared, and Depth Point Cloud Data Fusion 
        & Grassland Monitoring and Protection 
        & \checkmark  
        & $\times$ 
        & \cite{zhang2024grassland} \\
        \hline
        $\bullet$ Laplacian Pyramid Transform (LPT) \newline
        $\bullet$ Fuzzy Logic
        & RGB and Thermal Images 
        & Fruit Detection 
        & \checkmark  
        & \checkmark  
        & \cite{bulanon2009fruitdetection} \\
        \hline
        $\bullet$ Backpropagation Neural Network \newline
        $\bullet$ Independent Recurrent Neural Network
        & Meteorological and Area Data 
        & Rice Crop Yield Prediction 
        & \checkmark  
        & \checkmark  
        & \cite{chu2020riceyield} \\
        \hline
        $\bullet$ Super-Resolution Convolutional Neural Network (ESRCNN)
        & Landsat-8 and Sentinel-2 Data 
        & Sunflower Lodging Identification
        & \checkmark  
        & $\times$ 
        & \cite{song2020sunflowerlodging} \\
        \hline
    \end{tabular}
\end{table*}

\section{Statistical and Bibliometric Analysis}  
\label{Sec 5}

Addressing \textbf{RQ-3}, this section outlines the findings of a statistical analysis conducted to evaluate research activity in ML for agriculture. The collected data indicate a substantial growth in research output in recent years, underscoring the dynamic and vibrant nature of this domain.

Fig.~\ref{research} illustrates the annual volume of publications at the intersection of ML and agriculture. The search criteria, which included the terms `machine learning' and `agriculture' in the title, abstract, or keywords, produced distinct yearly counts. A notable surge is observed between 2018 and 2020. An exponential trendline—with an $R^2$ value of 0.993—demonstrates an exceptional fit to the data, reflecting a robust correlation between publication count and time. This near-perfect correlation reinforces the rapidly escalating research momentum in ML applications for agriculture.


\begin{figure}
    \centering
    \begin{tikzpicture}
        \begin{axis}[
            ybar,
            bar width=12pt,
            width=14cm, height=8cm,
            xlabel={\textbf{\textcolor{black}{Year}}},
            ylabel={\textbf{\textcolor{black}{Publications}}},
            xlabel style={yshift=-10pt},  
            ylabel style={yshift=16pt},  
            xtick={2012,2013,2014,2015,2016,2017,2018,2019,2020,2021,2022,2023,2024,2025}, 
            xticklabels={,{\textbf{\textcolor{black}{2013}}},{\textbf{\textcolor{black}{2014}}},{\textbf{\textcolor{black}{2015}}},
                         {\textbf{\textcolor{black}{2016}}},{\textbf{\textcolor{black}{2017}}},{\textbf{\textcolor{black}{2018}}},
                         {\textbf{\textcolor{black}{2019}}},{\textbf{\textcolor{black}{2020}}},{\textbf{\textcolor{black}{2021}}},
                         {\textbf{\textcolor{black}{2022}}},{\textbf{\textcolor{black}{2023}}},{\textbf{\textcolor{black}{2024}}},}, 
            xticklabel style={rotate=45, anchor=east, text=black, font=\bfseries}, 
            ymin=0, ymax=250000,
            ytick={50000,100000,150000,200000,250000},  
            yticklabels={\textbf{\textcolor{black}{50000}}, \textbf{\textcolor{black}{100000}}, 
                         \textbf{\textcolor{black}{150000}}, \textbf{\textcolor{black}{200000}}, 
                         \textbf{\textcolor{black}{250000}}}, 
            scaled y ticks=false,  
            legend style={at={(0.5,1.15)}, anchor=north, legend columns=-1, 
                         text=black, font=\bfseries, draw=gray!50}, 
            axis x line=bottom, 
            axis y line=left,
            ymajorgrids=true,
            grid style={line width=0.3pt, draw=gray!30},
            enlarge x limits={0.1},  
            tick label style={black, font=\bfseries},  
            label style={black, font=\bfseries},  
            title style={black, font=\bfseries}  
        ]

        \addplot[fill=blue, draw=black] coordinates {
            (2013,25114) (2014,27420) (2015,32648) (2016,37384)
            (2017,46202) (2018,60263) (2019,87369) (2020,105550)
            (2021,131781) (2022,154472) (2023,184773) (2024,223425)
        };

        \addplot[domain=2013:2024, samples=100, smooth, ultra thick, red] 
            {22444 * exp(0.212 * (x-2013))};

        \addlegendimage{ybar, fill=blue, draw=black}
        \addlegendentry{\textbf{\textcolor{black}{Publications}}}
        
        \addlegendimage{legend image code/.code={
            \draw[very thick, red] (0cm, 0cm) -- (0.7cm, 0cm); 
        }}
        \addlegendentry{\textbf{\textcolor{black}{Trend Line \boldmath$22444e^{0.212x} \times R^2 = 0.993$}}}

        \end{axis}
    \end{tikzpicture}
    \caption{{The Amount of Research Carried Out in ML for Agriculture with Trend line. Data source: Scopus on December 30, 2024}}  
    \label{research}
\end{figure}

Table~\ref{publication-distribution} provides the distribution of publications by document type, along with their corresponding percentage shares. Articles comprise 49.63\% of the total, while conference papers account for 37.00\%. The remaining document types contribute smaller percentages, ranging from 0.20\% for editorials to 4.45\% for book chapters. This table succinctly captures the publication landscape within ML for agriculture.

\begin{figure}[htbp]
    \centering
    \includegraphics[width=0.99\textwidth]{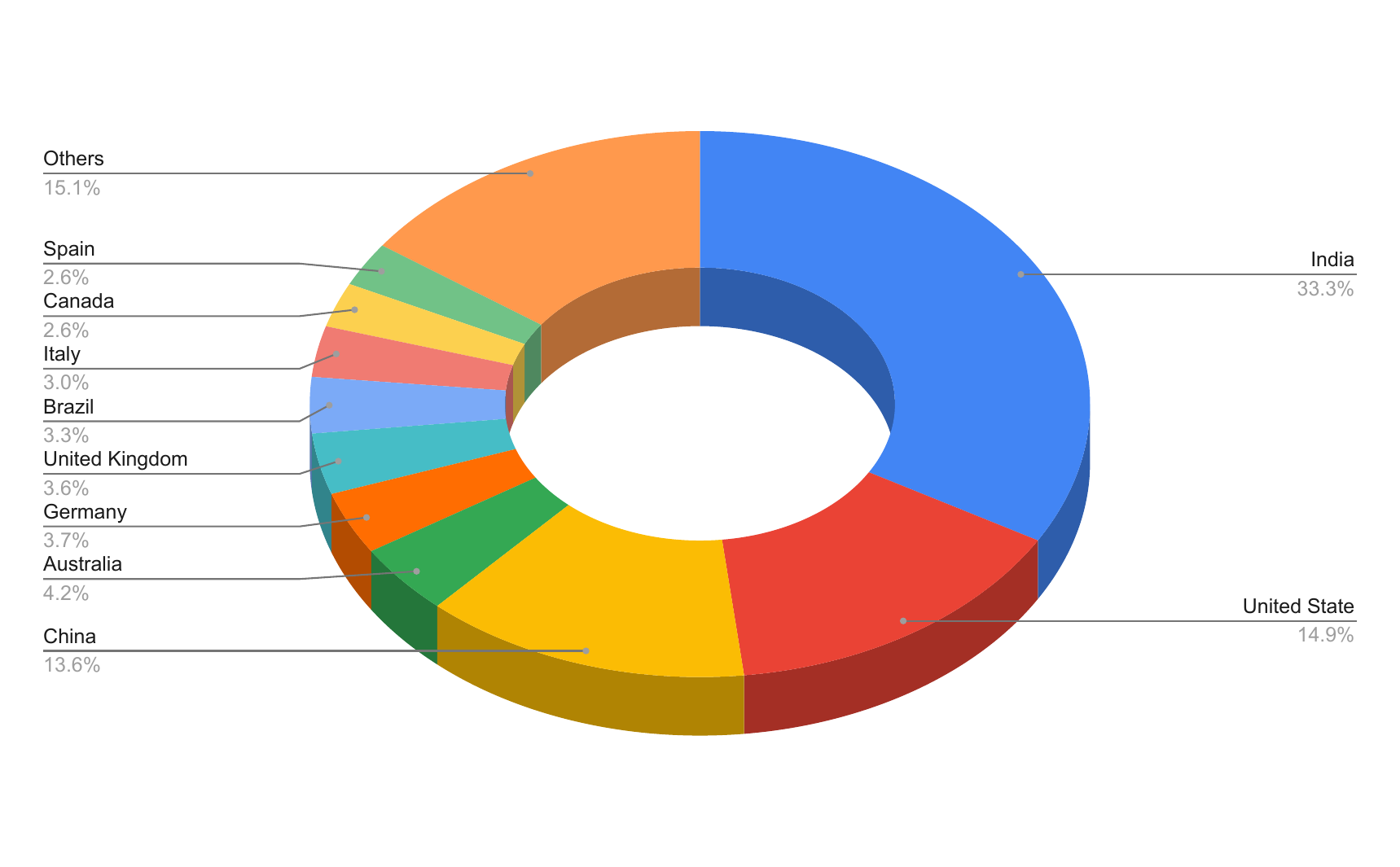}
    \caption{Research Contribution by Countries in ML for Agriculture. Data Source: Scopus on December 30, 2024}
    \label{contribution}
\end{figure}

Fig.~\ref{contribution} presents the research contributions by various countries. The pie chart clearly shows the proportional distribution of publications, with India leading, followed by the United States and China. This visual representation offers a concise overview of global research involvement in the field.

\begin{table}[htbp]
  \centering
  \caption{Publication Distribution by Document Type. Data Source: Scopus on December 30, 2024}
  \label{publication-distribution}
  \renewcommand{\arraystretch}{1.3}
  \begin{tabular}{|@{}l|l|r|@{}|}
    \toprule
    Document Type & No. of Publications & Percentage (\%) \\
    \hline
    Article & 3190 & 49.63 \\
    Conference Paper & 2378 & 37.00 \\
    Review & 362 & 5.63 \\
    Book Chapter & 286 & 4.45 \\
    Conference Review & 147 & 2.29 \\
    Book & 28 & 0.44 \\
    Data Paper & 23 & 0.36 \\
    Editorial & 13 & 0.20 \\
    \midrule
    Total & 6427 & 100\\
    \bottomrule
  \end{tabular}
\end{table}

\begin{figure}
    \centering
    \includegraphics[width=1.0\textwidth]{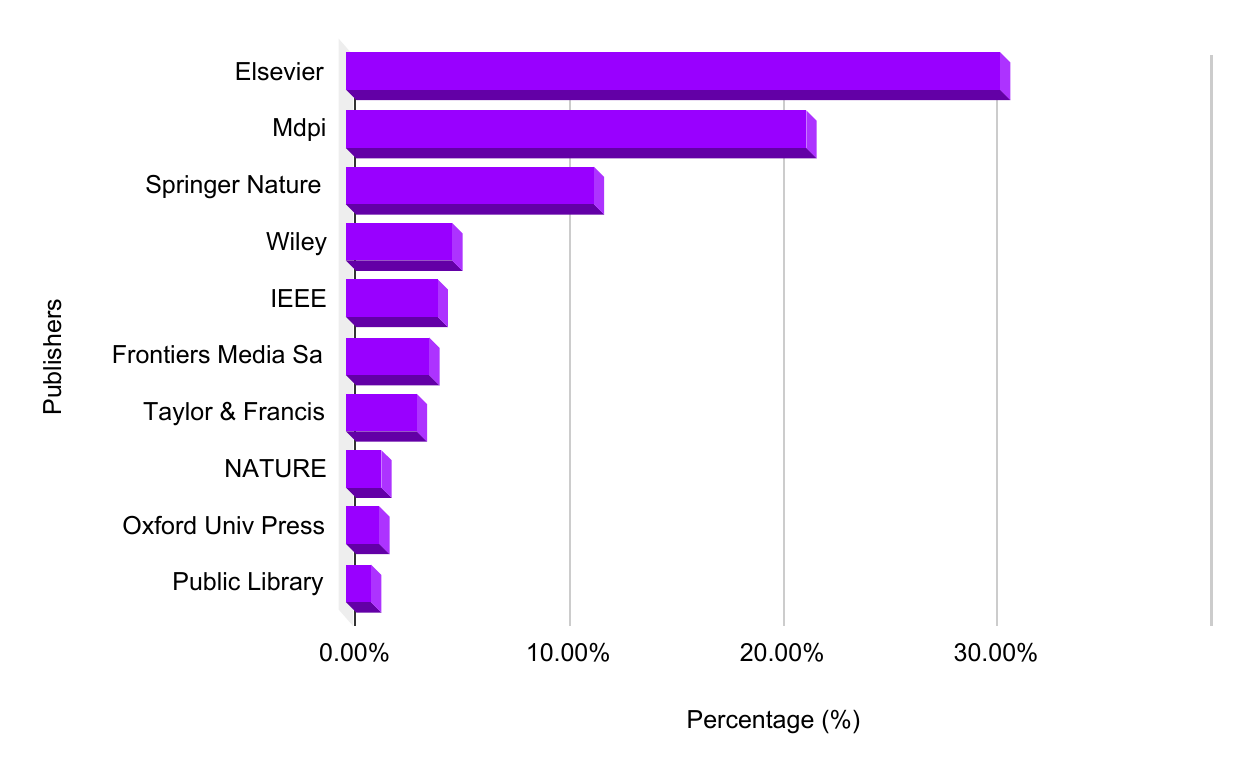}
    \caption{Number of Publications by Different Types of Publishers. Data Source: Scopus on December 30, 2024}
    \label{publi}
\end{figure}

Fig.~\ref{publi} depicts the number of publications by different types of publishers, showing that Elsevier is the leading publisher, followed by MDPI, Springer Nature, Wiley, IEEE, and others.

\begin{figure*}[htbp]
    \centering
    \includegraphics[width=1\textwidth]{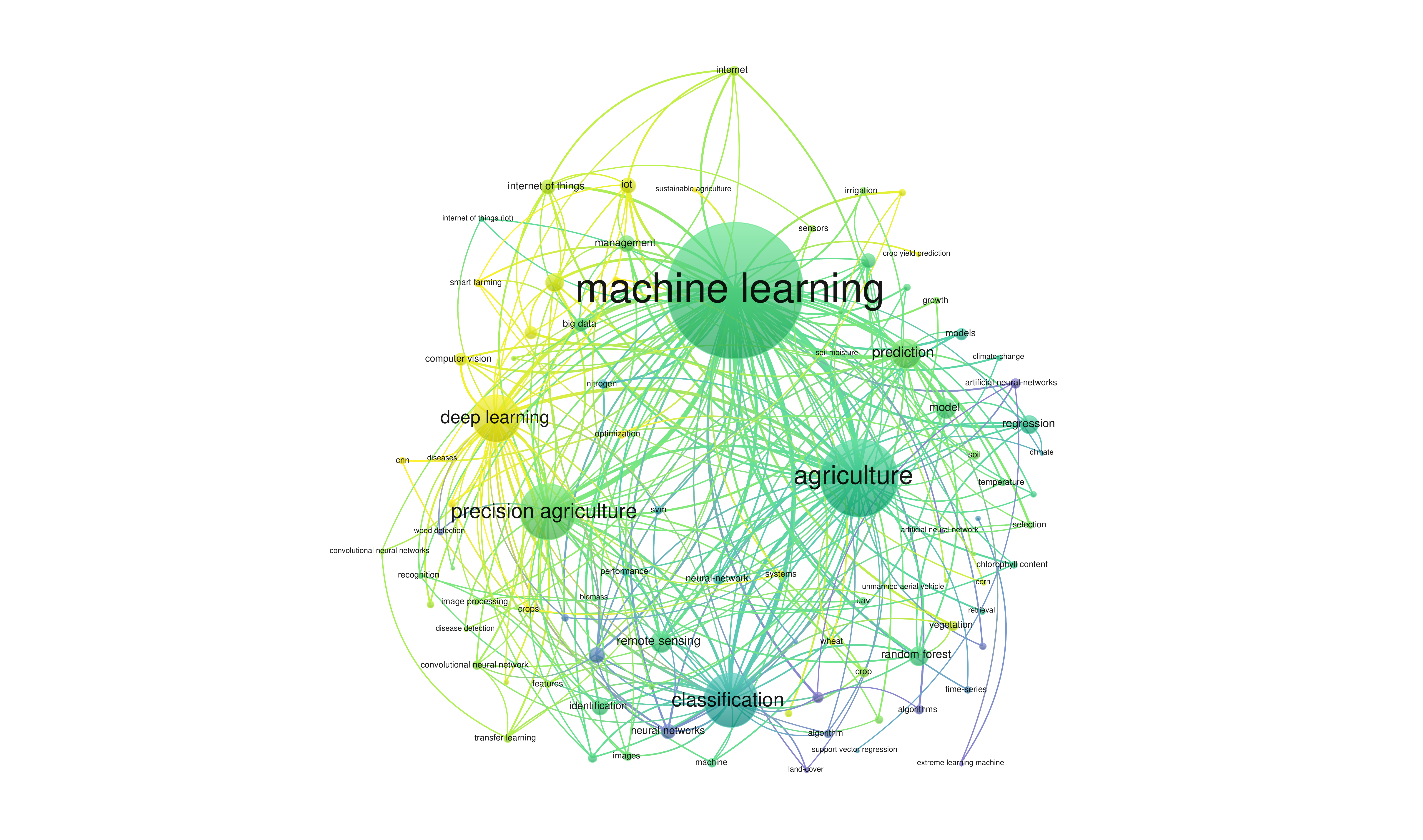}
    \caption{The Bibliometric Analysis of Keywords in ML Application in Agriculture. Data Source: Web of Science on December 30, 2024}
    \label{key}
\end{figure*}

Fig.~\ref{key} presents a bibliometric network of keywords in ML applications for agriculture, generated using VoSviewer \cite{van2010software}. Keywords such as `precision farming', `crop yield prediction', `plant disease detection', and `soil analysis' emerge as prominent, reflecting the central research themes in the field.

\begin{figure*}[htbp]
    \centering
    \includegraphics[width=1.0\textwidth]{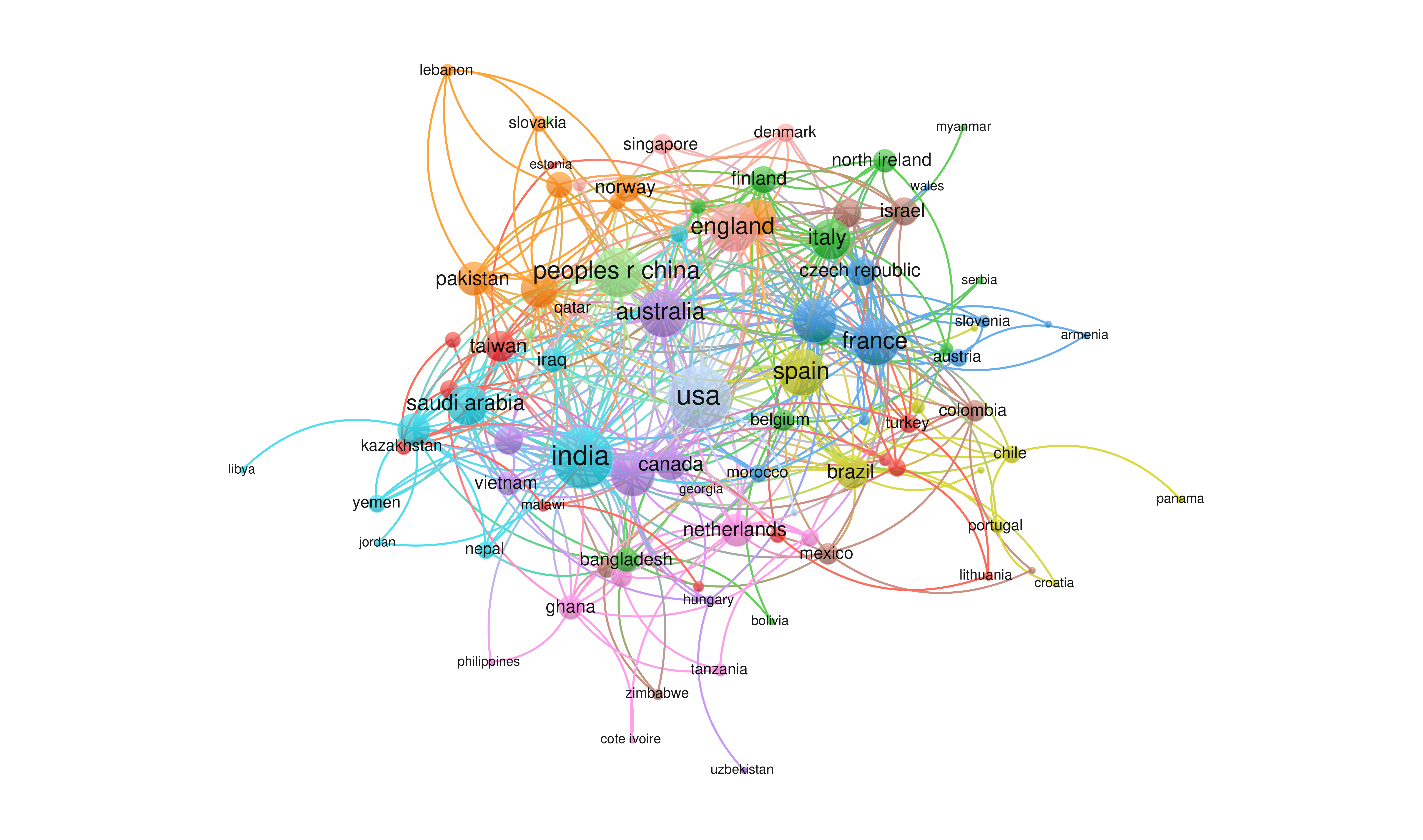}
    \caption{The bibliometric analysis of countries in ML application in agriculture. Data source: Web of Science on December 30, 2024}
    \label{countri}
\end{figure*}

Fig.~\ref{countri} offers a bibliometric mapping of countries contributing to ML research in agriculture. The analysis identifies India, the United States, and China as leading contributors, with additional nations grouped under "Others," underscoring the global reach of research in this area.

\begin{figure*}[htbp]
    \centering
    \includegraphics[width=1\textwidth]{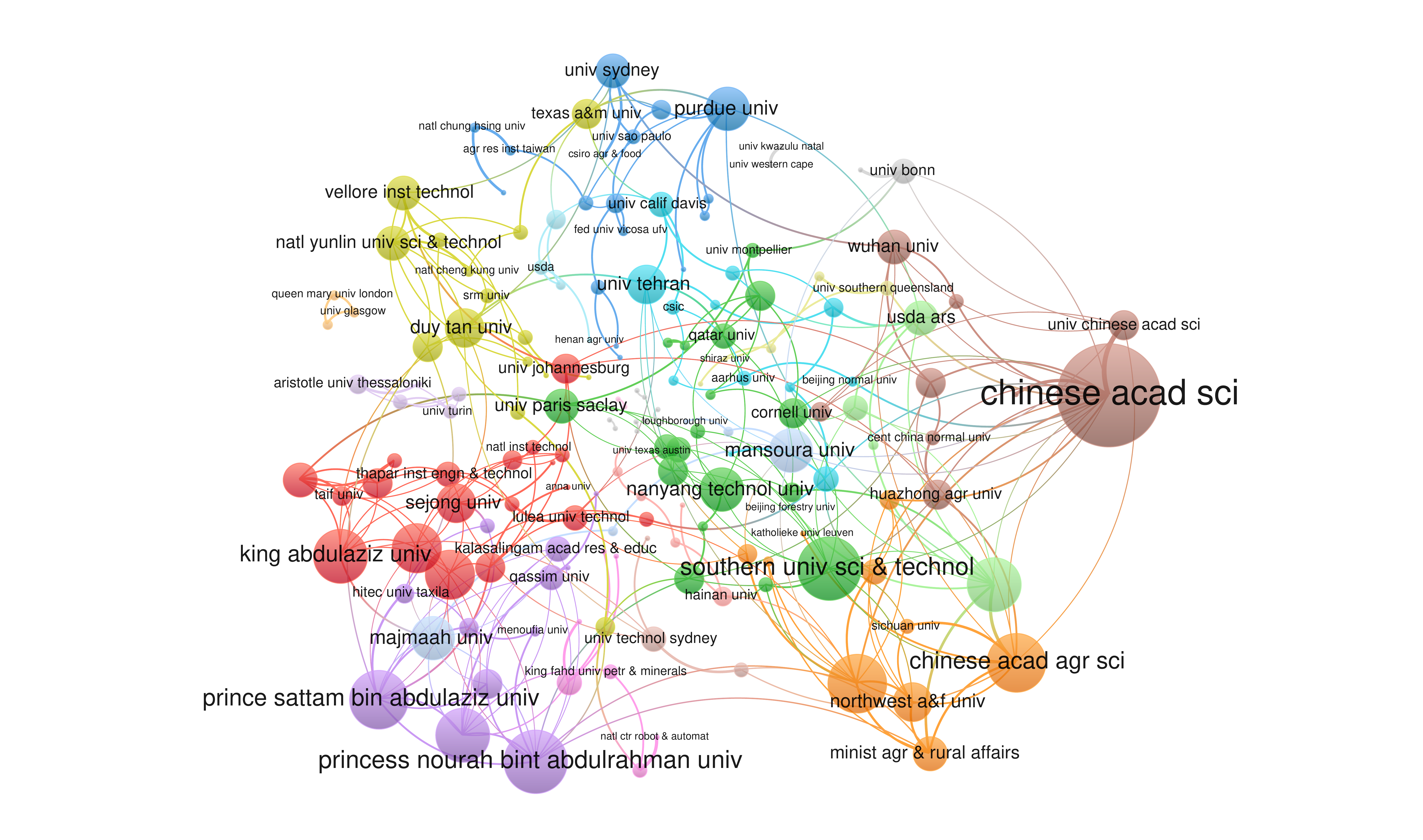}
    \caption{The Bibliometric Analysis of Organizations in ML Application in Agriculture. Data Source: Web of Science on December 30, 2024}
    \label{author}
\end{figure*}

Fig.~\ref{author} illustrates the bibliometric analysis of organizations in ML for agriculture. Institutions such as Aarhus University, Addis Ababa Science and Technology University, and Agricultural Research Center have emerged as key contributors, highlighting their impact and collaborative linkages within the research community.

Collectively, these statistical and bibliometric analyses provide a comprehensive overview of the research landscape in ML for agriculture. The rapid increase in publication volume, along with the diverse contributions from countries, publishers, and organizations, not only highlights current research trends but also lays a robust foundation for future collaborative efforts and advancements in the field.

\section{ Real-World Applications and Industry Trends}\label{Sec 6}
This section is dedicated to addressing \textbf{RQ-4} by reviewing real-world applications through case studies and discussing approximately ten AI and robotics companies in agriculture.
\subsection{Case Studies}

The use of AI in agriculture has enormous potential to revolutionize the way farming is conducted, from improving crop management to reducing costs and simplifying tasks for farmers. This potential is being realized by a number of start-ups and applications
 Here are some examples of how AI is being practically utilized (Table  \ref{tab:ai_agriculture}) in agriculture:

\begin{itemize}
   \item \textbf{Kisan GPT:} It launched on March 15, 2023, is an AI chatbot that caters to the underserved agriculture sector of India. It provides guidance on various topics such as irrigation, pest control, and crop cultivation. Unlike other popular AI chatbots where users are required to type their queries, Kisan GPT offers the option of voice input using a microphone. It supports 10 languages including Hindi, English, Gujarati, and Marathi, as well as foreign languages like Japanese, Portuguese, and Indonesian. With India being an agricultural powerhouse, this AI chatbot could be a game-changer for the agriculture sector. Farmer-friendly and accessible, it bridges the information gap between farmers and experts, aiding decision-making and enhancing crop management. With AI-powered chatbots, like Kisan GPT, farmers can achieve greater efficiency and make informed choices, benefiting India's agricultural prowess \cite{L8}.

    \item \textbf{Jiva:} Jiva was created by Olam International, a global agribusiness headquartered in Singapore. While it has operations in India too. Jiva provides a range of services, including micro-financing and AI-based systems, empowering farmers to unlock the potential of AI. With Jiva, farmers can enhance crop health, streamline operations, and increase profitability, ushering in a promising future for agriculture. AI powered systems like Jiva offer personalized recommendations for optimal farming practices, resulting in higher crop yields and improved profitability \cite{L9}.
    
    \item \textbf{See and Spray:} It is a US based system introduced by Blue River Technology. The 'See and Spray' system uses computer vision to identify individual plants and selectively apply herbicides, reducing the amount of chemicals needed and improving crop yields \cite{L10}.
    
    \item \textbf{xFarm:} xFarm Technologies is a Swiss-Italian company headquartered in Switzerland, with strong operations in Italy. xFarm is an AI-powered app that helps farmers optimize their irrigation and fertilizer usage. The app uses ML algorithms to analyze data from various sources, such as soil moisture sensors, weather data, and satellite imagery. With this information, it generates recommendations for farmers like when, how much water and fertilizer to apply to their crops. The app can also provide alerts if there are any issues with irrigation, such as leaks or blockages. By optimizing irrigation and fertilizer usage, the xFarm app helps farmers improve crop yields while reducing water and fertilizer waste. \cite{L11}
   
\end{itemize}

\begin{table}[ht]
\centering
\caption{Analystical analysis of AI-driven Agricultural Solutions in Case Studies}
\label{tab:ai_agriculture}
\resizebox{\textwidth}{!}{%
\begin{tabular}{|p{3cm}|p{4cm}|p{4.5cm}|p{4.5cm}|}
\hline
\textbf{Case Study Name} & \textbf{Data Sources Integrated} & \textbf{AI/ML Techniques Used} & \textbf{Impact on Agriculture and Farmers} \\ \hline

\textbf{Kisan GPT} & 
Weather data, expert recommendations, government advisories, farmer queries & 
$\bullet$ Reinforcement Learning from Human Feedback (RHFL)\newline
$\bullet$ LLMs\newline
$\bullet$ NLP\newline
$\bullet$ Speech Recognition & 
Real-time multilingual chatbot, removing the barrier of reading and language for farmers, aiding decision-making with expert-backed insights. \\ \hline

\textbf{Jiva} & 
Crop yields, soil health, financial records, farm productivity reports & 
$\bullet$ Predictive Analytics\newline
$\bullet$ Decision Trees\newline
$\bullet$ Financial Risk Modeling\newline
$\bullet$ AI sensors & 
Provides AI-based farming and micro-financing recommendations, improving financial sustainability. Offers free personalized agronomic advice, cash advances for farm inputs, and facilitates high-quality input purchasing and crop sales. \\ \hline

\textbf{See and Spray} & 
High-resolution camera images, GPS mapping, historical weed infestation patterns, weed data & 
$\bullet$ Deep Learning\newline
$\bullet$ Computer Vision\newline
$\bullet$ Sensor Fusion & 
Selective herbicide spraying, reducing chemical usage, improving weed control, and increasing crop yield. \\ \hline

\textbf{xFarm} & 
Soil moisture sensors, weather forecasts, satellite imagery & 
$\bullet$ ML-Based Predictive Models\newline
$\bullet$ Geospatial Analytics\newline
$\bullet$ Anomaly Detection & 
Optimized irrigation and fertilizer management, reducing resource waste and improving yield through data-driven recommendations. \\ \hline

\end{tabular}%
}
\end{table}

\subsection{Top 10 AI and Robotics Companies/Start-Up Disrupting Agriculture}

The use of AI and robotics in agriculture has gained momentum in recent years, and some companies have emerged as global leaders in this field. The top 10 global companies using AI and robotics in agriculture, according to their 2023 valuation, are Benson Hill Biosystems, Iron Ox, FarmWise, Carbon Robotics, CropIn, Blue River Technology, Aerobotics, Prospera, AgroScout, and Root AI. The details of this company are provided in Table \ref{tab:agriculture-companies}. These companies have developed various technologies, including ML, computer vision, and robotics, to optimize crop growth, monitor plant health, and detect diseases early. Using these technologies, they help farmers make data-driven decisions and achieve higher yields while reducing their environmental footprint. In general, the emergence of these companies is a testament to how AI and robotics can revolutionize agriculture and contribute to a more sustainable and efficient food system.

\begin{landscape} 
{\scriptsize
\begin{longtable}{|p{1cm}|p{2.5cm}|p{2.5cm}|p{1.5cm}|p{3cm}|p{3.5cm}|p{3.5cm}|p{3.5cm}|} 
\caption{Top 10 Global Companies Using AI and Robotics in Agriculture Based on Their 2023 Valuation.}
\label{tab:agriculture-companies} \\
\hline
\textbf{Serial No.} & \textbf{Company} & \textbf{Headquarters} & \textbf{Year Founded} & \textbf{AI/ML Techniques Used} & \textbf{Data Sources Integrated} & \textbf{Impact on Agriculture} & \textbf{Website} \\ 
\hline
\endfirsthead

\multicolumn{8}{c}{{\bfseries Continued from previous page}} \\
\hline
\textbf{Serial No.} & \textbf{Company} & \textbf{Headquarters} & \textbf{Year Founded} & \textbf{AI/ML Techniques Used} & \textbf{Data Sources Integrated} & \textbf{Impact on Agriculture} & \textbf{Website} \\ 
\hline
\endhead

\hline
\endfoot
\bottomrule
\endlastfoot

1 & Benson Hill Biosystems & USA & 2012 & 
$\bullet$ Machine Learning (ML) for crop genome analysis 
& Genetic sequencing data, soil properties, crop traits 
& Enhances crop breeding efficiency through AI-driven genetic insights. 
& \url{https://bensonhillbio.com/} \\ \hline

2 & Iron Ox & USA & 2015 & 
$\bullet$ AI-powered robotics \newline
$\bullet$ Reinforcement Learning 
& Environmental sensors, soil moisture, plant health data 
& Reduces water and fertilizer use while optimizing plant growth. 
& \url{https://www.ironox.com} \\ \hline

3 & FarmWise & USA & 2016 & 
$\bullet$ Computer Vision \newline
$\bullet$ ML for robotic weeding 
& Camera images, soil condition data, weed growth patterns 
& Lowers herbicide use by automating precision weeding. 
& \url{https://farmwise.io/} \\ \hline

4 & Carbon Robotics & USA & 2018 & 
$\bullet$ Deep Learning \newline
$\bullet$ AI-driven autonomous weeding robots 
& Multi-spectral imaging, GPS farm mapping 
& Eliminates weeds without chemicals, improving sustainability. 
& \url{https://carbonrobotics.com/} \\ \hline

5 & Blue River Technology & USA & 2011 & 
$\bullet$ Computer Vision \newline
$\bullet$ AI-based precision agriculture 
& High-resolution imaging, real-time plant analysis 
& Reduces chemical use with plant-specific treatments. 
& \url{http://bluerivertechnology.com} \\ \hline

6 & Root AI & USA & 2017 & 
$\bullet$ AI-powered autonomous robots for harvesting 
& Sensor-based crop monitoring, fruit ripeness detection 
& Improves farm efficiency using robotic automation. 
& \url{https://root-ai.com/} \\ \hline

7 & CropIn & India & 2010 & 
$\bullet$ AI-driven satellite imaging \newline
$\bullet$ Predictive Analytics 
& Remote sensing, weather forecasts, soil health data 
& Supports precision farming through risk management and remote crop monitoring. 
& \url{http://www.cropin.com/} \\ \hline

8 & Aerobotics & South Africa & 2014 & 
$\bullet$ Deep Learning for aerial imaging 
& Drone-captured images, temperature data, vegetation indices 
& Detects crop diseases early, minimizing yield losses. 
& \url{https://www.aerobotics.com/} \\ \hline

9 & Prospera & Israel & 2014 & 
$\bullet$ ML \newline
$\bullet$ Computer Vision for crop monitoring 
& Farm camera networks, soil health analytics 
& Enhances yield with real-time crop health analysis. 
& \url{https://www.prospera.ag/} \\ \hline

10 & AgroScout & Israel & 2017 & 
$\bullet$ Multi-Sensor Data Fusion \newline
$\bullet$ Deep Learning 
& Satellite imagery, IoT sensor data, weather patterns 
& Enables early-stage pest and disease detection. 
& \url{http://agro-scout.com/} \\ \hline

\end{longtable}
}
\end{landscape}

\section{Publicly available dataset related to Agriculture}
\label{Sec 7}

Training an ML model for experiments is highly dependent on the quality and structure of the dataset employed. Frequently, these datasets comprise raw data that must be preprocessed before model training. In addressing \textbf{RQ-5}, this section highlights several publicly available datasets to assist the research community in selecting appropriate data for their studies. These datasets are classified into three categories—weed control, fruit detection, and miscellaneous—as detailed in Table \ref{table:weed_control}, Table \ref{table:fruit_detection}, and Table \ref{table:miscellaneous_datasets}, respectively.

The available information for these datasets is summarized in terms of Modality and Annotation.\texttt{ Modality} refers to the different types of data available, such as multispectral, RGB, etc. \texttt{Annotation} refers to the task of labeling images. Specifically, the \texttt{ pixel-level annotation} \cite{sun2018image} also known as semantic segmentation, is the leveling of each pixle of an image with a specific class. This annotation provides a detailed understanding of the image by identifying the exact boundaries and shapes of the objects. 
On the other hand, \texttt{image level annotation} \cite{pinheiro2015image}, which is defined as labeling the entire image with a single class or a set of classes. This type of annotation gives an understanding of an image's content without the exact location of objects.


\begin{table*}[htb!]
\centering
\caption{Public Datasets Dedicated to Weed Control.}
\begin{tabular}{|p{0.5cm}|p{4.90cm}|p{3.75cm}|p{2.0cm}|p{1cm}|}
\toprule
\midrule
\textbf{No.} & \textbf{Dataset Name} & \textbf{Modality} & \textbf{Annotation} & \textbf{Ref.} \\ \hline
1 & CWFI dataset & Multispectral & Pixel level & \cite{cwfi_dataset} \\ \hline
2 & Carrot-Weed & RGB & Pixel level & \cite{carrot_weed} \\ \hline
3 & Plant seedlings & RGB & Image level & \cite{plant_seedlings} \\ \hline
4 & Grass-Broadleaf & RGB & Patch level & \cite{grass_broadleaf} \\ \hline
5 & Sugar Beets 2016 & Multimodal & Image level & \cite{sugar_beets_2016} \\ \hline
6 & Synthetic SugarBeet Weeds & RGB & Pixel level & \cite{synthetic_sugarbeet_weeds} \\ \hline
7 & WeedNet & Multispectral & Image level & \cite{weednet} \\ \hline
8 & Joint stem detection & Multispectral+RGB & Pixel level & \cite{joint_stem_detection} \\ \hline
9 & Leaf counting & RGB & Image level & \cite{leaf_counting} \\ \hline
10 & Weed Map & Multispectral & Pixel level & \cite{weed_map} \\ \hline
11 & DeepWeeds & RGB & Image level & \cite{deepweeds} \\ \hline
12 & Crop weed discrimination & Multispectral & Pixel level & \cite{crop_weed_discrimination} \\ \hline
13 & Early crop weed & RGB & Image level & \cite{early_crop_weed} \\ \hline
14 & Ladybird Cobbitty Brassica & Multimodal & No annotations & \cite{ladybird_cobbitty} \\ \hline
15 & Open Plant Phenotype Database (OPPD) & RGB & Image level, bounding box & \cite{oppd} \\ 
\midrule
\bottomrule
\end{tabular}
\label{table:weed_control}
\end{table*}


\begin{table*}[htb!]
\centering
\caption{Public Datasets Dedicated to Fruit Detection.}
\begin{tabular}{|p{1cm}|p{4.0cm}|p{2.5cm}|p{4cm}|p{1cm}|}
\toprule
\midrule
\textbf{No.} & \textbf{Dataset} & \textbf{Modality} & \textbf{Annotation} & \textbf{Ref.} \\ \hline
1 & DeepFruits & RGB & Bounding box & \cite{deepfruits} \\ \hline
2 & Orchard Fruit & RGB & Bounding box, circle & \cite{orchard_fruit} \\ \hline
3 & Date Fruit & RGB & Image level & \cite{date_fruit} \\ \hline
4 & KFuji RGB-DS & RGB-D & Bounding box & \cite{kfuji_rgbds} \\ \hline
5 & MangoNet & RGB & Pixel level & \cite{mangonet} \\ \hline
6 & MangoYOLO & RGB & Bounding box & \cite{mangoyolo} \\ \hline
7 & WSU apple dataset & RGB & Bounding box & \cite{wsu_apple} \\ \hline
8 & Fuji-SfM & RGB & Bounding box & \cite{fuji_sfm} \\ \hline
9 & LFuji-air dataset & LiDAR & Bounding box & \cite{lfuji_air} \\ \hline
10 & MinneApple & RGB & Pixel level & \cite{minneapple} \\ 
\midrule
\bottomrule
\end{tabular}
\label{table:fruit_detection}
\end{table*}

\begin{table*}[htb!]
\centering
\caption{Miscellaneous Public Datasets.}
\begin{tabular}{|p{1cm}|p{4.5cm}|p{3.0cm}|p{3.0cm}|p{1cm}|}
\toprule
\midrule
\textbf{No.} & \textbf{Dataset} & \textbf{Modality} & \textbf{Annotation} & \textbf{Ref.} \\ \hline
1 & 3D Broccoli & RGBD & No annotation & \cite{3d_broccoli} \\ \hline
2 & Apple Trees & RGBD & No annotation & \cite{apple_trees} \\ \hline
3 & Capsicum Annuum & RGB & Pixel level & \cite{capsicum_annuum} \\ \hline
4 & Fruit flower dataset & RGB & Pixel level & \cite{fruit_flower} \\ \hline
5 & Sugarcane billets & RGB & Image level & \cite{sugarcane_billets} \\ \hline
6 & Maize disease & RGB & Line level & \cite{maize_disease} \\ \hline
7 & DeepSeedling & RGB & Bounding box & \cite{deepseedling} \\ \hline
8 & GrassClover & RGB & Pixel level & \cite{grassclover} \\ \hline
9 & Oil radish growth & RGB & Pixel level & \cite{oil_radish_growth} \\ \hline
10 & University of Arkansas, Plants Dataset & Info not available & Info not available & \cite{arkansas_plants_dataset} \\ \hline
11 & EPFL, Plant Village Dataset & Info not available & Info not available & \cite{plant_village} \\ \hline
12 & Leafsnap Dataset & Info not available & Info not available & \cite{leafsnap} \\ \hline
13 & LifeCLEF Dataset & Info not available & Info not available & \cite{lifeclef} \\ \hline
14 & PASCAL Visual Object Classes Dataset & Info not available & Info not available & \cite{pascal_voc} \\ \hline
15 & Africa Soil Information Service (AFSIS) dataset & Info not available & Info not available & \cite{afsis_dataset} \\ \hline
16 & UC Merced Land Use Dataset & Info not available & Info not available & \cite{uc_merced_landuse} \\ \hline
17 & MalayaKew Dataset & Info not available & Info not available & \cite{malayakew} \\ \hline
18 & Crop/Weed Field Image Dataset & Info not available & Info not available & \cite{crop_weed_field_image} \\ \hline
19 & University of Bonn & Info not available & Info not available & \cite{university_bonn} \\ \hline
20 & Flavia Leaf Dataset & Info not available & Info not available & \cite{flavia_leaf} \\ \hline
21 & Syngenta Crop Challenge 2017 & Info not available & Info not available & \cite{syngenta_crop_challenge} \\ 
\midrule
\bottomrule
\end{tabular}
\label{table:miscellaneous_datasets}
\end{table*}

\section{Challenges and Future Research Directions}\label{Sec 8}

\subsection{Challenges and Limitations of Machine Learning in Agriculture} 

Although ML and IoT applications \cite{misra2020iot} in smart agriculture are changing the way work is done in the agricultural sector, there are still some limitations to the implementation of AI in agriculture that require further research. Few challenges and limitations include as follows:

    \begin{itemize}
        \item \textbf{Subsidies and welfare schemes:} This is a major issue that affects the agricultural sector in many countries. Despite the government's efforts to provide subsidies and welfare schemes, farmers often do not receive the benefits due to bureaucratic hurdles, corruption, and lack of awareness. This has a significant impact on the livelihoods of farmers and their ability to sustain their farms \cite{L5}.
        \item \textbf{Lack of information about water management:} Water is a critical resource for agriculture, and farmers need to manage it effectively to ensure optimal crop yield. However, many farmers lack access to information about water management techniques and technologies, such as drip irrigation, water harvesting, and soil moisture sensors. This can lead to inefficient water use and reduced crop yields \cite{L2,L7}.
        \item \textbf{Absence of agricultural marketing:} Agricultural marketing is essential for farmers to sell their produce at a fair price and reach a wider market. However, many farmers in rural areas lack access to proper markets and transportation facilities, which makes it difficult for them to sell their produce. This can result in low prices for farmers and food shortages for consumers \cite{L1,L3}.
        \item \textbf{Lack of proper storage facilities:} Proper storage facilities are essential for preserving the quality of agricultural produce and preventing post-harvest losses. However, many farmers lack access to such facilities, which can lead to spoilage and wastage of crops. This can also result in lower prices for farmers, as they are forced to sell their produce immediately after harvest \cite{L3}.
        \item \textbf{Limited knowledge of modern agricultural practices:} Agriculture is a rapidly evolving sector, with new technologies and practices being developed all the time. However, many farmers lack access to information about these new practices, which can result in low crop yields and reduced profitability. This also makes it difficult for farmers to compete in a globalized market, where other farmers may have access to the latest \cite{L4,ryan2022social}.
        \item \textbf{Lack of awareness about ICT benefits:} Information and communication technologies (ICT) can be powerful tools for farmers, allowing them to access information about weather patterns, market prices, and new technologies. However, many farmers are not aware of the benefits of these technologies or lack access to them. This can result in missed opportunities for farmers to improve their yields and profitability \cite{ryan2022social,vern2022digital}.
    \end{itemize}

The agricultural industry and farmers are deeply concerned with sustainability, and any implementation of AI technology must be capable of not only meeting but exceeding existing sustainability standards \cite{ryan2019ethics}. However, the use of AI may also lead to less sustainable practices and outcomes if not implemented properly. One potential consequence of relying on AI in agriculture is the possibility of an increase in pesticide usage, potentially leading to the adoption of more potent and hazardous pesticides without human oversight. Furthermore, if human labor is substituted with large, weighty robots, this could exacerbate preexisting issues linked to soil compaction, a problem already associated with the use of heavy machinery in farming \cite{carolan2020automated}.

The safe deployment of AI poses a significant challenge that cannot be overlooked. Farmers have raised concerns that the use of AI-powered robots  \cite{wakchaure2023application} and drones in agriculture could result in increased risks and harm to the environment, as there may be less attention given to potential harm to humans. This underscores the importance of carefully considering all potential consequences of AI implementation in farming and taking appropriate measures to mitigate any negative impacts.

Trust building within the agricultural sector is quite challenging as well \cite{pylianidis2021introducing}. When autonomous machines are deployed on farms and collect data that is transmitted back to agribusinesses, it becomes challenging to establish trust, particularly since farmers typically have limited knowledge about how their data is being utilized. In the agricultural industry, data privacy represents a significant concern, as farmers may need to compromise their privacy to reap the benefits and potential advantages of AI technology on their farms. Consequently, there is a growing apprehension about the possibility of AI infringing on farmers' privacy \cite{stock2021make}. 

Additionally, the ownership and control of the data retrieved, stored, and used by the AI system \cite{guo2019ifusion} is another issue that needs to be addressed.
Some other challenges and limitations of AI in agriculture include:

\begin{itemize}
    \item \textbf{Limited data availability:} Limited data availability for training AI models, due to the lack of technology infrastructure and limited connectivity in rural areas, is a major challenge for implementing AI in agriculture \cite{L6}.
    \item \textbf{Crop and soil variability:} Crop and soil variability, which makes it challenging to create a `unique' model that works for all conditions, requires AI models to be tailored to specific local conditions.
    \item \textbf{Weather variability:} Weather variability, which makes it difficult to predict crop yields and adapt to changing conditions, requires AI models to be constantly updated and refined to account for changing weather patterns.
    \item \textbf{High cost of technology:} The high cost of technology required for implementing AI in agriculture, such as sensors, drones, and other hardware, can be a barrier to adoption for small-scale farmers.

    \item \textbf{Bridging the Awareness Gap:} Limited awareness among farmers about new technologies and ethical concerns surrounding AI in agriculture is another barrier to adoption.
\end{itemize}

While AI has the potential to revolutionize the agricultural sector, there are still many challenges and limitations that need to be addressed before widespread adoption can occur. Further research is needed to overcome these challenges and ensure that the benefits of AI in agriculture are fully realized.

\subsection{Challenges in Data Fusion for Agricultural Applications/Heterogeneous Data Fusion Challenges}

As mentioned in the dataset section, various types of data are used in the agriculture field, originating from multiple sources such as RGB images from cameras, remote sensing, UAV images, satellite imagery, IoT sensors, weather models, soil quality reports, and economic datasets. The term data fusion refers to the process of combining data from multiple sources to produce more accurate, consistent, and concise information than that provided by individual sources \cite{munir2021ai}. While machine learning (ML) applications have advanced precision agriculture, data fusion challenges remain a significant barrier to effective decision-making.

\begin{itemize}

 \item\textbf{Data Format Incompatibility:}  
Agricultural data is available in multiple formats, making data fusion complex. For instance, satellite, drone, and camera images are stored as raster formats (JPEG, PNG, etc.), whereas meteorological data is represented as time-series or numerical data (JSON, SQL, CSV, etc.). Combining these diverse formats requires specialized techniques to align their structure and ensure compatibility for analysis \cite{mena2024adaptive, smith2006approaches}.
    
   \item\textbf{Spatial and Temporal Misalignment:}  
Different data sources operate at different time scales and resolutions. For instance, data from IoT sensors is collected in real time, while satellite images are taken approximately every 5 to 10 days. Therefore, fusing these types of data is impractical without preprocessing and aligning them to the same time scale \cite{maillet2023fusion}
     
     \item\textbf{Computational Complexity:}
Working with deep learning models on high-dimensional and large-scale datasets often requires GPUs.  For instance, fusing time-series data with spatial pixel-based data for agricultural applications can be achieved using a hybrid architecture that combines CNN and LSTM models, in this scenario, data fusion further increases computational complexity \cite{meng2020survey}.

 \item\textbf{Network Latency and Connectivity Issues:} 
    Smart farming practices require real-time decision-making, which heavily depends on IoT-based agricultural systems relying on wireless connections (Wi-Fi or strong internet connectivity). However, poor connectivity in rural areas leads to delayed data transmission, which, in turn, affects the real-time processing and predictions of ML models \cite{cravero2022challenges}.

\item\textbf{Missing and Conflicting Data from Multiple Sources:}  
One of the major challenges in applying ML models to smart farming is handling missing, inconsistencies, and conflicting data, all of which negatively effect real time  agricultural decision-making. Various factors contribute to these issues, including weather conditions, misplacement of IoT sensors, and other environmental influences. For example, satellite data may be unavailable due to cloud cover, resulting in missing data. Additionally, different data sources may provide inconsistent measurements—for instance, soil pH readings from IoT sensors might differ from lab-based data—making it difficult for ML models to accurately assess real conditions \cite{ouhami2021computer}.

\end{itemize}

\subsection{Future Directions of Machine Learning in Agriculture}
\label{Sec 10}

AI and AI-driven solutions are touted as the panacea of our age, and agriculture is one area where the benefits of AI applications are pragmatic and realistic. Here are some future directions of ML in agriculture:

\begin{itemize}
    \item \textbf{IoT standardization:} It is crucial to develop credibility and establish a market for a new concept, but the integration of physical objects into the internet poses several challenges regarding the compatibility of existing internet protocols and applications with these objects \cite{parasuraman2021iot}.
    
    \item \textbf{IoT and Multi sensors data :} The reliability, consistency, and volume of agricultural data may be compromised due to equipment malfunctions, network node failures, post-processing errors, or pest, disease outbreaks and data from multi sensors . These data gaps can negatively impact the accuracy of calculations and hinder the effectiveness of IoT sensors applications in agriculture \cite{haseeb2020energy}.
    
    \item \textbf{Regulatory issues:} It is imperative to establish clear regulatory and legal frameworks for the management and ownership of farm data, particularly with regards to the relationship between farm laborers and data companies. These regulations may vary from country to country, taking into account factors such as service delivery, technological complexity, market competition, and data security. This variance in regulatory practices across different regions can impact the application of IoT in specific agricultural contexts, such as monitoring and agro-food supply chains \cite{vern2022digital}.
        
    \item \textbf{Security:} Smart agriculture relies on IoT devices to operate effectively, but these devices are at risk of physical tampering that could impede their function. Common sources of tampering include theft, damage by livestock and rodents, relocation, and connectivity problems. Such tampering can result in a range of issues, including data loss, equipment damage, and increased expenses for farmers. As such, it is important to take steps to secure IoT devices and minimize their vulnerability to physical interference \cite{vangala2023security, saha2021iot}. 
    
    \item \textbf{Market issues:} Given the agricultural industry's tendency to operate with slim profit margins, it is important to carefully strategize the integration of IoT-enabled technologies by considering their potential long-term benefits \cite{haseeb2020energy, mokaya2019future}.
    
    \item \textbf{Drones and unmanned aerial vehicles:} Unmanned aerial vehicles (UAVs)
    \cite{su2023ai} offer a cost-effective and broad approach to environmental monitoring by capturing images and collecting data about specific scenes. As practical applications for drone technology continue to advance, the use of drones in agriculture is expected to expand in the coming years \cite{boursianis2020internet}.
    
    \item \textbf{Driverless tractors:} Robotic agriculture is on the horizon and is expected to materialize in the next decade or so. In this future, autonomous tractors will replace human labor and perform all aspects of farming. These tractors will be fitted with sensors that will allow them to monitor obstacles and apply necessary inputs. Additionally, they will be able to perform required practices, ultimately leading to increased efficiency and productivity in the agriculture sector \cite{mokaya2019future}.
    
    \item \textbf{Automated irrigation systems:} Automated irrigation systems, when combined with historical weather data, can accurately predict the necessary resources required for optimal crop growth. These systems utilize real-time ML to maintain the desired soil conditions, which leads to an increase in average yields. One of the most significant benefits of these systems is the reduction in labor required for irrigation. Additionally, they have the potential to reduce production costs. The system's reliance on historical weather data ensures that resources are used efficiently, which helps to reduce waste. Automated irrigation systems provide farmers with greater control over the irrigation process, resulting in healthier crops and higher yields \cite{mokaya2019future}.
    
    \item \textbf{Crop health monitoring:} Conventional crop health monitoring techniques are laborious and largely categorical. However, companies are now competing to develop hyperspectral imaging and 3D laser scanning technologies that can enhance precision and accuracy by collecting vast amounts of data. By using deep learning \cite{zhang2018survey,liu2018pixel} techniques, an alert-based system can be developed to improve crop protection \cite{mokaya2019future}.
    
    \item \textbf{Model Interpretation:} ML models are often considered black boxes, making it challenging for researchers and practitioners to understand why and how they make decisions \cite{shams2024enhancing}. These challenges raise concerns about the trustworthiness of their predictions in  agricultural applications for farmers. Developing interpretable models for agricultural applications can help mitigate these issues \cite{cartolano2024analyzing}.
    
    \item \textbf{Incorporating Generative Adversarial Networks:} Generative Adversarial Networks (GANs) can address data scarcity and imbalance in agricultural datasets for ML and deep learning models. Agricultural datasets often lack diversity in environmental conditions, leading to overfitting or underfitting in ML and DL models \cite{lu2022generative}. Beyond image generation, GANs offer solutions for multi-sensor and multi-source data fusion. Future research can explore domain-specific GAN architectures to enhance their applicability in agriculture \cite{wang2023review}.

\end{itemize}

\section{Discussion and Conclusion}
\label{Sec 9}

This review provides a comprehensive analysis of over 70 research articles on ML applications in agriculture, addressing the research objectives outlined at the beginning of the study. In doing so, it highlights key advancements and real-world implementations, such as Kisan GPT, Jiva, CropIn, and Blue River Technology, while offering a foundational resource for new researchers.

Addressing \textbf{RO-1}, the review investigates recent advancements and applications of ML across all agricultural stages—from pre-harvesting (e.g., soil classification, weather prediction, seed testing, and disease detection) to harvesting (e.g., fruit detection and robotic picking) and post-harvesting processes (e.g., quality assessment and food security). Tables \ref{tab:long} and \ref{breeding} succinctly summarize these contributions, providing a quick reference to key developments in each domain.

In line with \textbf{RO-2}, the study demonstrates how ML is integrated with agricultural data and data fusion techniques to extract valuable insights. It discusses various methodologies for data fusion, including ensemble learning, Bayesian inference, hybrid models, and deep learning-based feature fusion, that collectively enhance decision-making and productivity in precision agriculture.

Addressing \textbf{RO-3}, a detailed bibliometric analysis is presented, revealing a substantial growth in research output over recent years. Figures \ref{key}, \ref{countri}, and \ref{author} illustrate research trends, global contributions, and institutional collaborations. The statistical findings, supported by an exponential trendline (with an $R^2$ value of 0.993) and publication distribution data (Table \ref{publication-distribution}), underscore the dynamic and vibrant nature of ML research in agriculture.

In response to \textbf{RO-4}, real-world applications and case studies are reviewed to bridge the gap between academic research and practical implementation. The discussion of case studies and the analysis of approximately ten leading AI and robotics companies in agriculture (detailed in Table \ref{tab:ai_agriculture}) showcase how these technologies are revolutionizing farming practices on a global scale.

Finally, addressing \textbf{RO-5}, the review lists several publicly available datasets categorized into weed control, fruit detection, and miscellaneous datasets. These datasets, detailed in Tables \ref{table:weed_control}, \ref{table:fruit_detection}, and \ref{table:miscellaneous_datasets}, serve as a valuable resource for practitioners aiming to build ML models for agricultural applications.

Collectively, these findings not only highlight current research trends and technological advancements but also lay a robust foundation for future collaborative efforts and innovations in the application of ML in agriculture. While this review does not delve into the technical intricacies of individual ML algorithms or the broader socio-economic impacts of agricultural AI, it provides comprehensive insights that will guide future research directions and practical implementations in the field.

\section*{Acknowledgement}

The author expresses her gratitude to the Ministry of Human Resource Development (MHRD), Government of India, for providing financial assistance for her research and  ANID FONDECYT project 123112,  ANID PIA/PUENTE AFB230002, Chile.


\end{document}